\documentclass[10pt,twocolumn,letterpaper]{article}

\usepackage{cvpr}
\usepackage{times}
\usepackage{epsfig}
\usepackage{graphicx}
\usepackage{amsmath}
\usepackage{amssymb}
\usepackage{bm}
\usepackage{cite}
\usepackage[percent]{overpic}
\usepackage{multirow}
\usepackage[table]{xcolor}   

\def\bal#1\eal{\begin{align}#1\end{align}} 
\def\suml{\sum\limits}

\newcommand{\br}[1]{\left[#1\right]} 
\newcommand{\pr}[1]{\left(#1\right)} 
\newcommand{\cbr}[1]{\left\{#1\right\}} 
\DeclareMathOperator*{\argmin}{arg\,min} 
\def\transp{\mathsf{T}} 
\def\m{\mathbf}
\def\mc{\mathcal}
\def\R{\mathbb{R}}
\def\ast{*}
\def\opname{\operatorname}
\newcommand{\grad}{\ensuremath{\nabla}} 
\newcommand{\norm}[2]{\ensuremath{\left\|#1\right\|_{#2}}}
\newcommand{\abs}[1]{\ensuremath{\lvert #1\rvert}}
\newcommand {\bbmtx}{\begin{bmatrix}} 
\newcommand {\ebmtx}{\end{bmatrix}} 

\usepackage[pagebackref=true,breaklinks=true,letterpaper=true,colorlinks,bookmarks=false]{hyperref}

\cvprfinalcopy 

\def\cvprPaperID{0} 

\begin{document}

\title{Non-local Color Image Denoising with Convolutional Neural Networks}

\author{Stamatios Lefkimmiatis\\
Skolkovo Institute of Science and Technology (Skoltech), Moscow, Russia\\
{\tt\small s.lefkimmiatis@skoltech.ru}
}

\maketitle

\begin{abstract}
We propose a novel deep network architecture for grayscale and color image denoising that is based on a
non-local image model. Our motivation for the overall design of the proposed network stems from  
variational methods that exploit the inherent non-local self-similarity property of natural 
images. We build on this concept and introduce deep networks that perform non-local processing and 
at the same time they significantly benefit from discriminative learning. Experiments on the Berkeley segmentation dataset, comparing several state-of-the-art methods, show that the proposed non-local models achieve the best reported denoising performance both for grayscale and color images for all the tested noise levels. It is also worth noting that this increase in performance comes at no extra cost on the capacity of
the network compared to existing alternative deep network architectures. In addition, we highlight a direct link of the proposed non-local models to convolutional neural networks. This connection is of significant importance since it allows our models to take full advantage of the latest advances on GPU computing in deep learning and makes them amenable to efficient implementations through their inherent parallelism. 
\end{abstract}

\section{Introduction}
Deep learning methods have been successfully applied
in various computer vision tasks, including image classification~\cite{Krizhevsky2012,He2016}
and object detection~\cite{Erhan2014,Ren2015}, and have dramatically improved the performance of these systems,
setting the new state-of-the-art. Recently, very promising results have 
also been reported for image processing applications such as image restoration~\cite{Burger2012,Xie2012},  super-resolution~\cite{Kim2016}
and optical flow~\cite{Bailer2015}. 

The significant boost in performance achieved by deep networks can be mainly attributed to their advanced modeling capabilities, thanks  
to their deep structure and the presence of non-linearities that are combined with discriminative learning on large training datasets. However, most of the current deep learning methods developed for image restoration tasks are based on general network architectures that do not fully exploit problem-specific knowledge. It is thus reasonable to expect that incorporating such information could lead to further improvements in performance. Only very recently, Schmidt and Roth~\cite{Schmidt2014} and Chen and  Pock~\cite{Chen2016} introduced deep networks whose architecture is specifically tailored to certain image restoration problems. However, even in these cases, the resulting models are local ones and do not take into account the inherent \emph{non-local self-similarity} property of natural images. On the other hand, conventional methods that have exploited this property have been shown to gain significant improvements compared to standard local approaches. A notable example is the Block Matching and 3D Collaborative Filtering (BM3D) method~\cite{Dabov2007} which is a very efficient and highly engineered approach that held the state-of-the-art record in image denoising for almost a decade.

\begin{figure}[t]
\centering
\begin{tabular}{@{} c @{ } c @{} }
   \includegraphics[width=.5\linewidth]{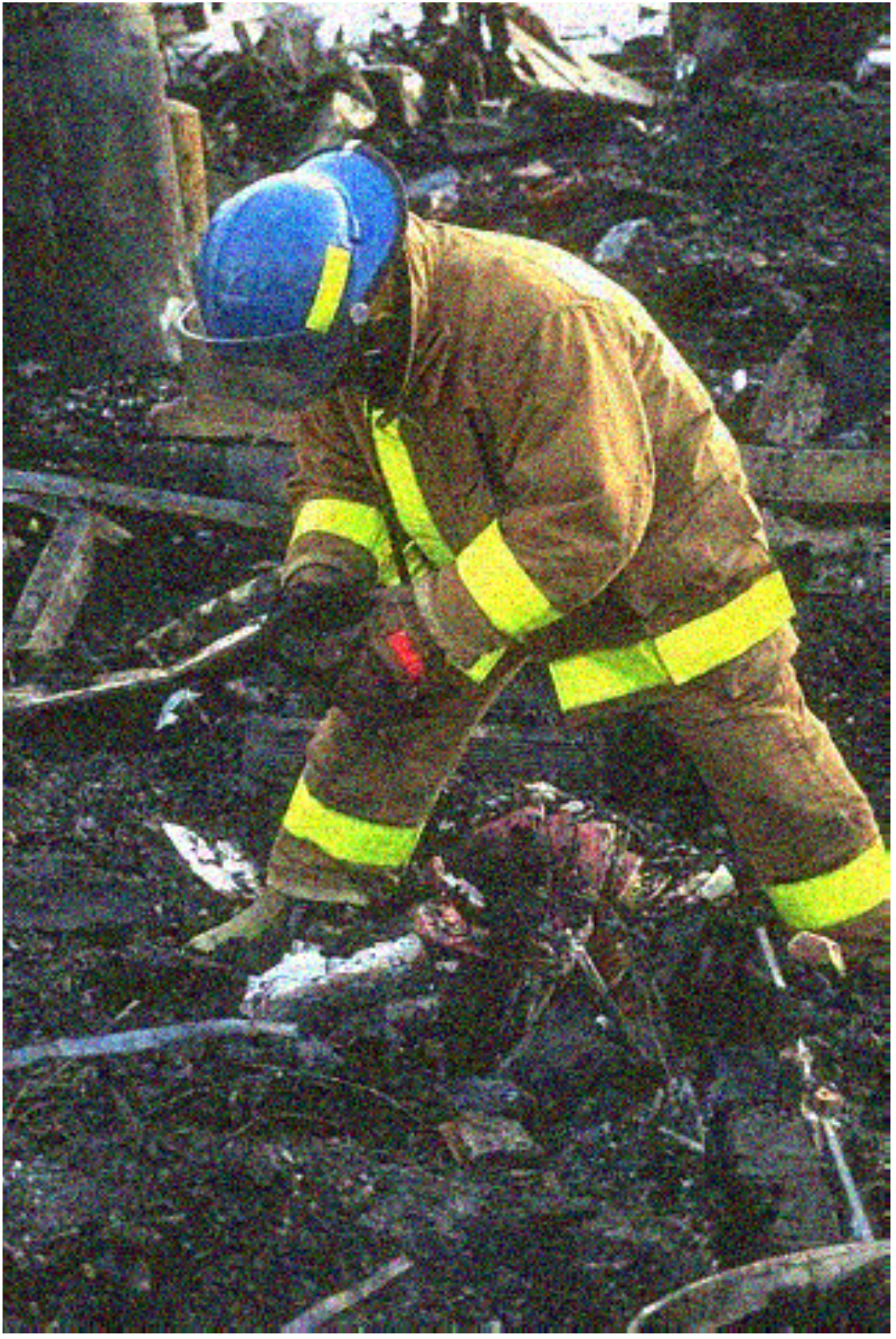}&
    \includegraphics[width=.5\linewidth]{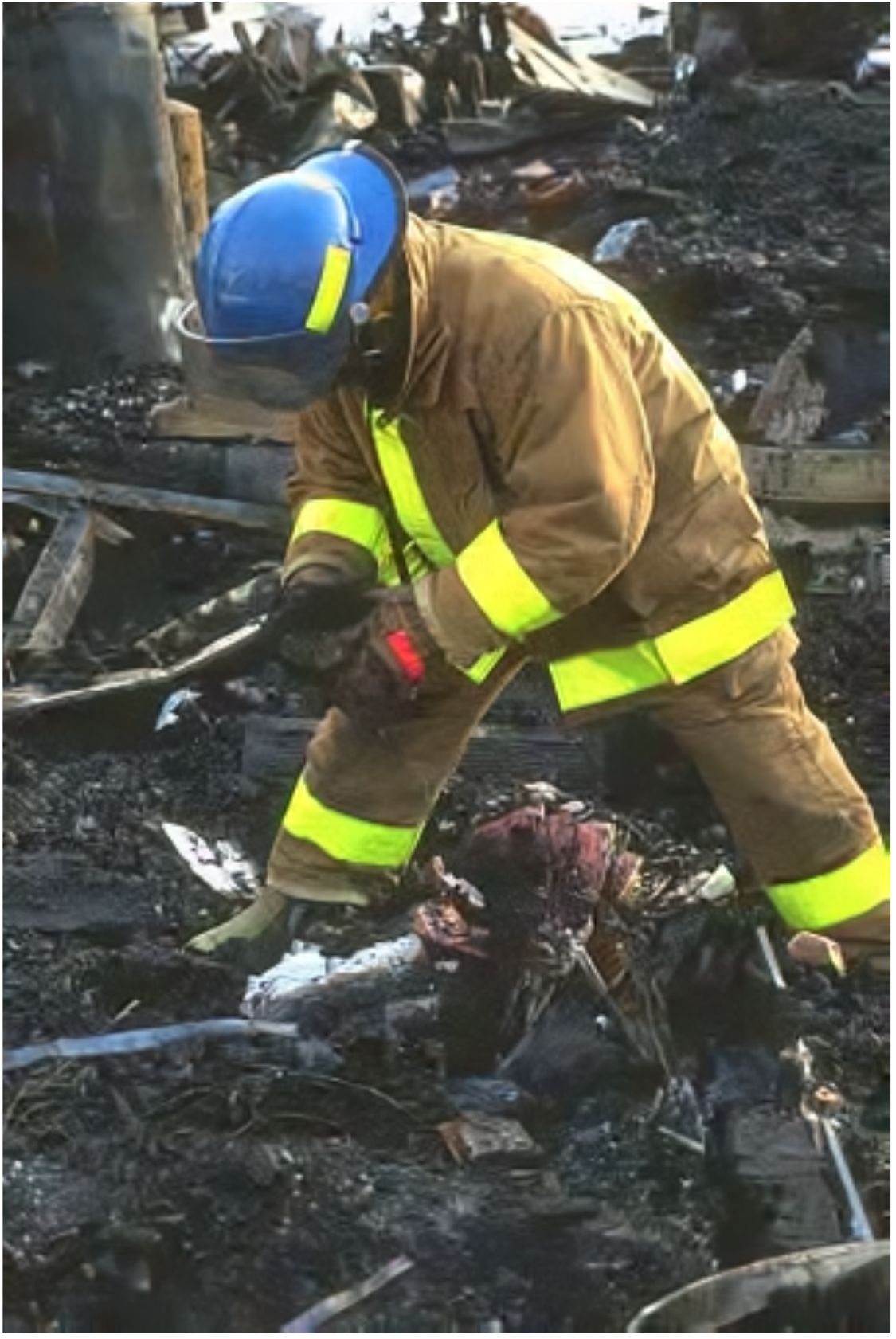} \\
    (a) & (b)
\end{tabular}
   \caption{Image denoising with the proposed deep non-local CNN model. (a) Noisy image corrupted with additive Gaussian noise ($\sigma = 25$) ; $\operatorname{PSNR} = 20.16 \text{ dB}$. (b) Denoised image using the 5-stage feed-forward network described in Sec.~\ref{sec:colorDen} ; $\operatorname{PSNR} = 29.53 \text{ dB}$. }
   \label{fig:CNLNet}
\end{figure}

In this work, motivated by the recent advances in deep learning and relying on the rich body of algorithmic ideas that
have been developed in the past for tackling image reconstruction problems, we study deep network architectures for image denoising. Inspired by non-local variational methods and other related approaches, we design a network that performs non-local processing and at the same time it significantly benefits from discriminative learning. Specifically, our strategy is instead of manually designing a non-local regularization functional, to learn the non-local regularization operator and the potential function following a loss-based training approach. 

Our contributions in this work can be summarized as follows:
\textbf{(1)} We propose a novel deep network architecture that is discriminatively trained for image denoising. 
As opposed to the existing deep-learning methods for image restoration, which are based on local models, our network explicitly models the non-local self-similarity property of natural images through a grouping operation of similar patches and a joint filtering. 
\textbf{(2)} We unroll a proximal gradient method into a deep network and learn the relevant parameters using a simple yet effective back-propagation strategy. 
\textbf{(3)} In contrast to the majority of recent denoising methods that are designed for processing single-channel images, we introduce a variation of our network that applies to color images and leads to state-of-the-art results. 
\textbf{(4)} We highlight a direct link of our proposed non-local networks with convolutional neural networks (CNNs). This connection allows our models to take full advantage of the latest advances on GPU computing in deep learning and makes them amenable to efficient implementations through their inherent parallelism.

\section{Variational Image Restoration Revisited}
The goal of image denoising is the restoration of a grayscale or color image $\m X$ from a corrupted observation $\m Y$, with the later obtained according to the observation model 
\bal
\m y = \m x + \m n\,.
\label{eq:forwardModel}
\eal
In this setting, $\m y$,  $\m x\in\R^{N\cdot C}$ are the vectorized versions of the observed and latent images, respectively,  $N$ is the number of pixels, $C$ the number of image channels, and $\m n$ is assumed to be i.i.d Gaussian noise with variance $\sigma^2$. 

Due to the \emph{ill-posedness} of the studied problem~\cite{Vogel2002}, Eq.~\eqref{eq:forwardModel} that relates the latent image to the observation cannot uniquely characterize the solution. This implies that in order to obtain a physically or statistically meaningful solution, the image evidence must be combined with suitable image priors.

Among the most popular and powerful strategies available in the literature for combining the observation and prior information is the variational approach. In this framework the recovery of $\m x$ from $\m y$ heavily relies on the formation of an objective function
\bal
E\pr{\m x} = D\pr{\m x, \m y} + \lambda J\pr{\m x},
\label{eq:ObjectiveFun}
\eal
whose role is to quantify the quality of the solution. Typically the objective function consists of two terms, namely the \emph{data fidelity term} $D\pr{\m x, \m y}$, which measures the proximity of the solution to the observation, and 
the \emph{regularizer} $J\pr{\m x}$ which constrains the set of plausible solutions by penalizing those that do not exhibit the desired properties. The regularization parameter $\lambda \ge 0$ balances the contributions of the two terms. Then, the restoration task is cast as the minimization of this objective function and the minimizer corresponds to the restored image. Note that for the problem under consideration and since the noise corrupting the observation is i.i.d Gaussian, the data term should be equal to ${\textstyle \frac{1}{2}}\norm{\m y - \m x}{2}^2$. This variational restoration approach has also direct links to Bayesian estimation methods and can be interpreted either as a penalized maximum likelihood or a maximum a posteriori (MAP) estimation problem~\cite{Bertero1998,Figueiredo2007}.

\subsection{Image Regularization}
\label{sec:ImReg}
The choice of an appropriate regularizer is very important, since it is one of the main factors that determine the quality of the restored image. For this reason, a lot of effort has been made to design novel regularization functionals that can model important image properties and consequently lead to improved reconstruction results. Most of the existing regularization methods are based either on a synthesis- or an analysis-based approach. Synthesis-based regularization takes place in a sparsifying-domain, such as the wavelet basis, and the restored image is obtained by applying an inverse transform~\cite{Figueiredo2007}. On the other hand, analysis-based regularization involves regularizers that are directly applied on the image one aims to restore. For general inverse problems, the latter regularization strategy has been reported to lead to better reconstruction results~\cite{Elad2007,Selesnick2009} and therefore is mostly preferred.

The analysis-based regularizers are typically defined as:
\bal
J\pr{\m x} = \suml_{r=1}^R \phi\pr{\m L_r \m x},
\eal
where $\m L: \R^N \mapsto \R^{R \times D}$ is the regularization operator ($\m L_r\m x$ denotes the D-dimensional $r$-th entry of the result obtained by applying $\m L$ to 
the image $\m x$) and $\phi : \R^D \mapsto \R$ is the potential function. Common choices for $\m L$ are differential operators of the first or of higher orders such as the gradient~\cite{Rudin1992,Bredies2010}, the structure tensor~\cite{Lefkimmiatis2015J}, the Laplacian and the Hessian~\cite{Lefkimmiatis2012J,Lefkimmiatis2013J}, or wavelet-like operators such as wavelets, curvelets and ridgelets~(see \cite{Figueiredo2007} and references therein). For the potential function $\phi$ the most popular choices are vector and matrix norms, but other type of functions are also frequently used such as the $\ell_0$ pseudo-norm and the logarithm.
Combinations of the above regularization operators and potential functions lead to existing regularization functionals that have been proven  very effective in several inverse problems, including image denoising.  A notable representative of the above regularizers is the Total Variation (TV)~\cite{Rudin1992}, where the regularization operator corresponds to the gradient and the potential function to the $\ell_2$ vector norm.

TV regularization and similar methods that penalize derivatives are essentially local methods, since they involve operators that act on a restricted region of the image domain. More recently, a different regularization paradigm has been introduced where non-local operators are employed to define new regularization functionals~\cite{Zhou2005,Kindermann2005,Elmoataz2008,Gilboa2008,Lefkimmiatis2015bJ}. The resulting non-local methods are well-suited for image processing and computer-vision applications and produce very competitive results. The reason is that they allow long-range dependencies between image points and are able to exploit
the inherent \emph{non-local self-similarity} property of natural images. This property implies that images often consist of localized patterns
that tend to repeat themselves possibly at distant locations in the image domain.

It is worth noting that alternative image denoising methods that do not fall in the category of analysis-based regularization schemes but still exploit the self-similarity property have been developed and produce excellent results. A non-exhaustive list of these methods is the non-local means filter (NLM)~\cite{Buades2010}, BM3D~\cite{Dabov2007}, the Learned Simultaneous Sparse Coding (LSSC)~\cite{Mairal2009},  and the Weighted Nuclear Norm Minimization (WNNM)~\cite{Gu2014}.

\subsection{Objective Function Minimization}
Besides the formulation of the objective function and the proper selection of the regularizer, another important aspect in the variational approach is the minimization strategy that will be employed to obtain the solution. For the case under study, the solution to the image denoising problem can be mathematically formulated as:
\bal
\m x^{\ast}&=\argmin_{a\le x_n \le b} \frac{1}{2}\norm{\m y - \m x}{2}^2 + \lambda \suml_{r=1}^{R}\phi\pr{\m L_r \m x} \nonumber\\
&=\argmin_{\m x} \frac{1}{2}\norm{\m y - \m x}{2}^2 + \lambda \suml_{r=1}^{R}\phi\pr{\m L_r \m x} + \iota_{\mc{C}}\pr{\m x}
\label{eq:energyMin}
\eal
where $\iota_{\mc{C}}$ is the indicator function 
of the convex set $\mc{C} = \cbr{\m x \in \R^N | x_n \in \left[ a, b\right] \forall n =1, \hdots N}$. The indicator function $\iota_{\mc{C}}$ takes the value 0 if $\m x \in \mc{C}$ and $+\infty$ otherwise. The presence of this additional term in Eq.~\eqref{eq:energyMin} stems from the fact that these type of constraints on the image intensities arise naturally. For example 
it is reasonable to require that the intensity of the restored image should either be non-negative (non-negativity constraint with $a=0, b = +\infty$) 
or its values should lie in a specific range (box-constraint). 


\subsection{Proximal Gradient Method}
There is a variety of powerful optimization strategies for dealing with Eq.~\eqref{eq:energyMin}. The simplest approach however, which we will follow in this work, is to directly use a gradient-descent algorithm. Since the indicator function $\iota_{\mc{C}}$ is non-smooth, instead of the classical gradient descent algorithm we employ the proximal gradient method~\cite{Parikh2013}. According to this method, the objective function is split into two terms, one of which is differentiable. Here we assume that the potential function $\phi$ is smooth and therefore we can compute its partial derivatives. In this case, the splitting that we choose for the objective function has the form $E\pr{\m x} = f\pr{\m x} + \iota_{\mc{C}}\pr{\m x}$, where $f\pr{\m x}$ is defined as 
\bal
f\pr{\m x} = \frac{1}{2}\norm{\m y - \m x}{2}^2 + \lambda \suml_{r=1}^{R}\suml_{d=1}^D \phi_d\pr{\pr{\m L_r \m x}_d}.
\eal
Note that in the above definition we have gone one step further and we have expressed the multivariable potential function $\phi$ as the sum of $D$ single-variable functions,  
\bal
\phi\pr{\m z} =\sum_{d=1}^D \phi_d\pr{\m z_d}.
\label{eq:decoupledPotentialFun}
\eal
As it will become clear later, this choice will allows us to reduce significantly the computational cost for training our network and will make the learning process feasible. It is also worth noting that this decoupled formulation of the potential function is met frequently in image regularization, as in wavelet regularization~\cite{Figueiredo2007}, anisotropic TV~\cite{Esedoglu2004} and Field-of-Experts (FoE)~\cite{Roth2009}.

\begin{figure*}[!t]
\centering
   \includegraphics[width=1\linewidth]{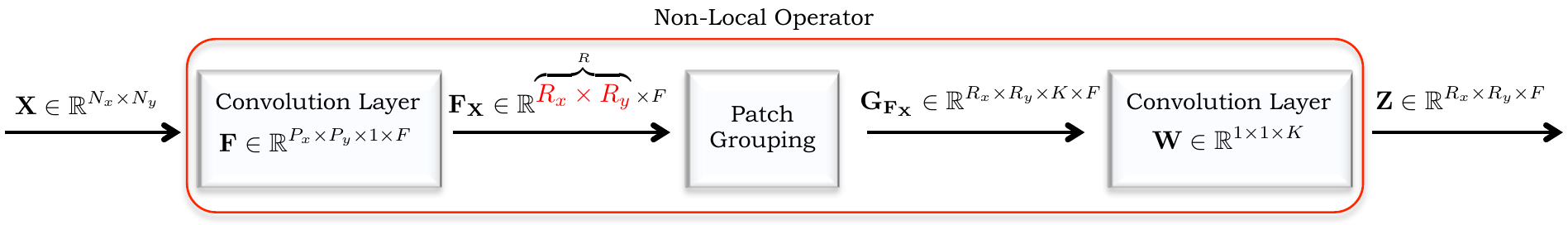}
   \caption{Convolutional implementation of the non-local operator of Eq.~\eqref{eq:NLOperator}.}
   \label{fig:NLOperator}
\end{figure*}

After the splitting of the objective function, the proximal gradient method recovers the solution in an iterative fashion, using the updates
\bal
\m x^{t} & =\mbox{prox}_{\gamma^t\iota_C}\pr{\m x^{t-1} - \gamma^t \grad_{\m x} f\pr{\m x^{t-1}}},
\eal
where $\gamma^t$ is a step size and $\mbox{prox}_{\gamma^t\iota_C}$ is the proximal operator~\cite{Parikh2013} related to the indicator function $\iota_{\mc{C}}$. The proximal map in this case corresponds to the orthogonal projection of the input onto $\mc{C}$, and hereafter will be denoted as $P_{\mc{C}}$. 

Given that the gradient of $f$ is computed as 
\bal
\grad_{\m x} f\pr{\m x} = \m x - \m y + \lambda \suml_{r=1}^R\m L_r^{\transp}\psi\pr{\m L_r \m x},
\eal 
where 
$\psi\pr{\m z} = \bbmtx \psi_1\pr{\m{z}_1} & \psi_2\pr{\m{z}_2}& \hdots & \psi_D\pr{\m{z}_D}\ebmtx^{\transp}$
and $\psi_d\pr{z}=\frac{d\phi_d}{d z}\pr{z}$, each proximal gradient iteration can be finally re-written as
\bal
\hspace{-.2cm}
\m x^{t}\! &=\!P_{\mc{C}}\!\pr{\!\m x^{t-1}\pr{1\!-\!\gamma^t} + \gamma^t \m y\!-\!\alpha^t\!\suml_{r=1}^R\m L_r^{\transp}\psi\pr{\m L_r \m x^{t-1}}\!\!}\!\!,
\label{eq:proxIter}
\eal 
where $\alpha^t = \lambda\gamma^t$.

In order to obtain the solution of the minimization problem in Eq.~\eqref{eq:energyMin} using this iterative scheme, a large number of iterations is required. In addition, the exact form of the operator $\m L$ and the potential function $\phi$ must be specified. Determining appropriate values for these quantities is in general a very difficult task. This has generated increased research interest and a lot of effort has been made for designing regularization functionals that can lead to good reconstruction results. 

\section{Proposed Non-Local Network}
\label{sec:NLNet}

In this work, we pursue a different approach than conventional regularization methods and instead of hand-picking the exact forms of the potential function and the regularization operator, we design a network that has the capacity to learn these quantities directly from training data. The core idea is to unroll the proximal gradient method and use a limited number of the iterations derived in Eq.~\eqref{eq:proxIter} to construct the graph of the network. Then, we learn the relevant parameters by training the network using pairs of corrupted and ground-truth data.  

Next, we describe in detail the overall architecture of the proposed network, which is trained discriminatively for image denoising. First we motivate and derive its structure for processing grayscale images, and then we explain the necessary modifications for processing color images. 

\subsection{Non-Local Regularization Operator}
\label{sec:NLOperator}
As mentioned earlier, non-local regularization methods have been shown to produce superior reconstruction results than their local counterparts~\cite{Gilboa2008,Lefkimmiatis2015bJ} for several inverse problems, including image denoising. Their superiority in performance is mainly attributed to their ability of modeling complex image structures by allowing long-range dependencies between points in the image domain. This fact highly motivates us to explore the design of a network that will exhibit a similar behavior. To this end, our starting point is the definition of a non-local operator that will serve as the backbone of our network structure.  

Let us consider a single-channel image $\m{X}$ of size $N_x \times N_y$
and let $\m x\in \R^N$, where $N = N_x \cdot N_y$, be the vector that is formed by stacking together 
the columns of $\m X$. Further, we consider image patches of size $P_{x} \times P_y$ and we denote 
by $\m x_r\in\R^P$,  with $P = P_x \cdot P_y$, the vector whose elements correspond to the 
pixels of the $r$-th image patch extracted from $\m X$. The vector $\m x_r$ is derived
from $\m x$ as $\m x_r = \m P_r \m x$, where $\m P_r$ is a $P \times N$ binary matrix that indicates which elements of $\m x$
belong to $\m x_r$. For each one of the $R$ extracted image patches, its $K$ closest neighbors are selected.  
Let $i_r = \cbr{i_{r,1},i_{r,2},\hdots,i_{r,K}}$, with $r = 1,\hdots,R$, be the set of indices of the $K$ most similar patches to the $r$-th patch $\m x_r$\footnote{The convention used here is that the set $i_r$ includes the reference patch, \ie $i_{r,1} = r$.}.
Next, a two-dimensional transform is applied to every patch $\m x_r$. The patch transform can be represented by a matrix-vector
multiplication $\m f_r = \m F\m x_r$ where $\m F \in \R^{F\times P}$. Note that if $F > P$ then the patch representation in the transform domain is redundant. In this work, we focus on the non-redundant case where $F=P$. For the transformed patch $\m f_r$, 
a group is formed using the $K$-closest patches. This is denoted as 
\bal
\m f_{i_r} = \bbmtx \m f_{i_{r,1}}^{\transp} & \m f_{i_{r,2}}^{\transp}	
& \hdots & \m f_{i_{r,K}}^{\transp}\ebmtx^{\transp}\in\R^{F\cdot K}.
\eal 
The final step of the non-local operator involves collaborating filtering among the group, which can be expressed as ${\m z}_r = \m W \m f_{i_r}$, where $\m W\in\R^{F \times \pr{F\cdot K}}$
is a weighting matrix and is constructed by retaining the first $F$ rows of a circulant matrix. The first row of this matrix corresponds to the vector $\m r =\bbmtx \bm w_1 & \hdots & \bm w_K\ebmtx\in\R^{F\cdot K}$, where $\bm w_i = \bbmtx w_i & 0 &\hdots & 0\ebmtx\in\R^F$. This collaborative filtering amounts to performing a weighted sum of the $K$ transformed patches in the group, \ie 
\bal
\m z_r = \suml_{k=1}^K w_k \m f_{i_{r,k}}.
\label{eq:NLSum}
\eal

\begin{figure*}[t]
\hspace{-0.5cm}
   \includegraphics[width=1.08\linewidth]{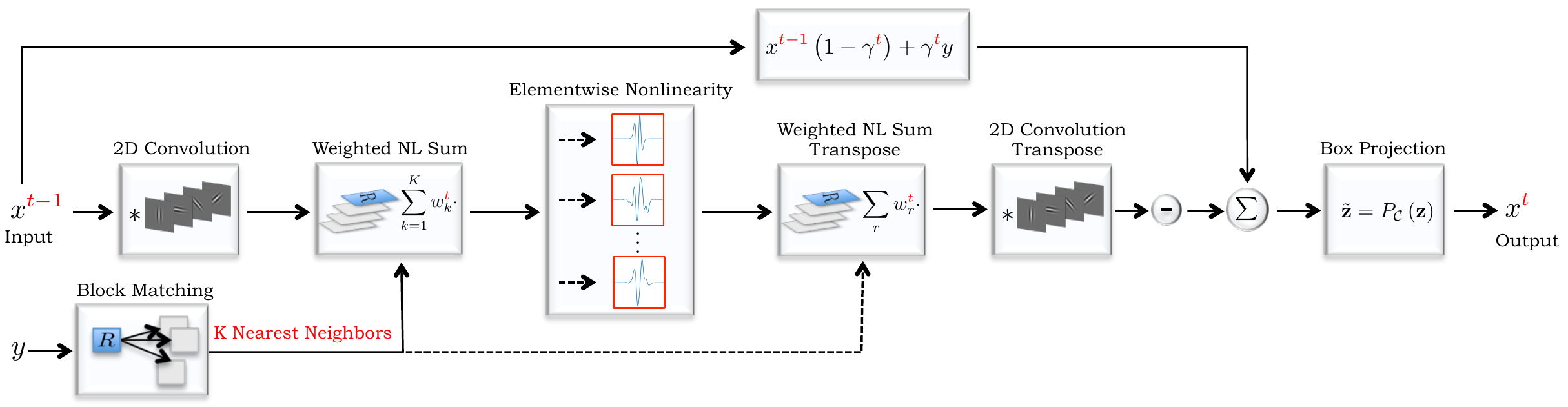}
   \caption{Architecture of a single stage of the proposed non-local convolutional network. Each stage of the network is symmetric and consists of both convolutional and de-convolutional layers. In between of these layers there is a layer of trainable non-linear functions.}
\label{fig:NLNet}
\end{figure*}

Based on the above, the non-local operator acting on an image patch $\m x_r$ can be expressed as the composition of three linear operators, that is
\bal
\operatorname{\m L}_r \m x = \pr{\m W \tilde{\m F} \m P_{i_r}}\m x,
\label{eq:NLOperator}
\eal
where $\m P_{i_r} = \bbmtx \m P_{i_{r,1}}^{\transp} & \m P_{i_{r,2}}^{\transp}& \hdots &\ \m P_{i_{r,K}}^{\transp}\ebmtx^{\transp}$ and $\tilde{\m F}\in\R^{\pr{F\cdot K}\times \pr{P\cdot K}}$ is a block diagonal matrix whose diagonal elements correspond to the patch-transform matrix $\m F$.
The non-local operator $\m L: \R^N \mapsto \R^{R\cdot F}$ described above bears strong resemblance to the BM3D analysis operator studied in~\cite{Danielyan2012}. The main difference between the two is that for the proposed operator in~\eqref{eq:NLOperator} a weighted average of the transformed patches in the group takes place, as described in Eq.~\eqref{eq:NLSum}, while for the operator of~\cite{Danielyan2012} a 1D Haar wavelet transform is applied on the group. Our decision for this particular set-up of the non-local operator was mainly based on computational considerations and for decreasing the memory requirements of the network that we propose next.

Due to the specific structure of the non-local operator $\m L_r$ (composition of linear operators) it  is now easy to derive its adjoint as 
\bal
\m L_r^{\transp} = \m P_{i_r}^{\transp}\tilde{\m F}^{\transp}\m W^{\transp}.
\label{eq:AdjointNLOperator}
\eal
The adjoint of the non-local operator is an important component of our network since it provides a reverse mapping from the transformed patch domain to the original image domain, that is $\m L^\transp: \R^{R\cdot F} \mapsto \R^N$.

\subsubsection{Convolutional Implementation of the Non-Local Operator}
As we explain next, both the non-local operator defined in~\eqref{eq:NLOperator} and its adjoint defined in~\eqref{eq:AdjointNLOperator} can be computed using convolution operations and their transpose. Therefore, they can be efficiently implemented using modern software libraries such as OMP and cuDNN that support multi-threaded CPU and parallel GPU implementations. 

Concretely, the image patch extraction and the 2D patch transform, $\m f_r = \m F\m P_r \m x$, can be combined and computed by passing the image $\m X$ from a convolutional layer. In order to obtain the desired output, the filterbank should consist of as many 2D filters as the number of coefficients in the transform domain. In addition, the support of these filters should match the size of the image patches. This implies that in our case $F$ filters with a support of $P_x \times P_y$ should be used. Also note that based on the desired overlap between consecutive image patches, an appropriate stride for the convolution layer should be chosen. Finally, the non-local weighted sum operation of~\eqref{eq:NLSum} can also be computed using convolutions. In particular, following the grouping operation of the similar transformed patches, which is completely defined by the set $I =\cbr{i_r : r =1\hdots R}$, the desired output can be obtained by convolving the grouped data with a single 3D filter of support $1 \times 1 \times K$. The necessary steps for computing the non-local operator using convolutional layers are illustrated in Fig.~\ref{fig:NLOperator}. To compute the adjoint of the non-local operator one simply has to follow the opposite direction of the graph shown in Fig.~\ref{fig:NLOperator} and replace the convolution and patch grouping operations with their transpose operations.

\subsection{Parameterization of the Potential Function}
Besides the non-local operator $\m L$, we further need to model the potential function $\phi$. We do this indirectly by representing its partial derivatives $\psi_i$ as a linear combination of Radial Basis Functions (RBFs),  that is 
\bal
\psi_i\pr{x} = \suml_{j=1}^M \pi_{ij}\rho_j\pr{\abs{x-\mu_j}},
\eal
where $\pi_{ij}$ are the expansion coefficients and $\mu_j$ are the centers of the basis functions $\rho_j$. There are a few radial functions to choose from~\cite{Hu2001}, but in this work we use Gaussian RBFs, $\rho_j\pr{r} = \exp\pr{-\varepsilon_j r^2}$. For our network we employ $M=63$ Gaussian kernels whose centers are distributed equidistantly and they all share the same precision parameter $\varepsilon$. 
The representation of $\psi_i$ using mixtures of RBFs is very powerful and allow us to approximate with high accuracy arbitrary non-linear functions. 
This is an important advantage over conventional regularization methods that mostly rely on a limited set of potential functions such as the ones reported in Section~\ref{sec:ImReg}. Also note that this parameterization of the potential gradient $\psi$ would have been computationally very expensive if we had not adopted the decoupled formulation of Eq.~\eqref{eq:decoupledPotentialFun} for the potential function.

Having all the pieces of the puzzle in order, the architecture of a single ``iteration" of our network, which we will refer to it as stage, is depicted in Fig.~\ref{fig:NLNet}. We note that our network follows very closely the proximal gradient iteration in Eq.~\eqref{eq:proxIter}. The only difference is that the parameter $\alpha^t$ has been absorbed by the potential gradient $\psi$, whose representation is learned. We further observe that every stage of the network consists of both convolutional and de-convolutional layers and in between there is a layer of trainable non-linear functions.

\subsection{Color Image Denoising}
\label{sec:colorDen}
The architecture of the proposed network as shown in Fig.~\ref{fig:NLNet} can only handle grayscale images. To deal with RGB color images, a simple approach would be to use the same network to process each image channel independently. However, this would result to a sub-optimal restoration performance since the network would not be able to explore the existing correlations between the different channels. 

To circumvent this limitation, we follow a similar strategy as in~\cite{Dabov2007} and before we feed the noisy color image to the network, we apply the same opponent color transformation which results to one luminance and two chrominance channels. Due to the nature of the color transform, the luminance channel contains most of the valuable information about primitive image structures and has a higher signal-to-noise-ratio (SNR) than the two chroma channels. We take advantage of this fact and since the block-matching operation can be sensitive to the presence of noise, we perform the grouping of the patches only from the luminance channel. Then, we use exactly the same set of group indices $I = \cbr{i_r : r = 1\hdots R}$ for the other two image channels. Another important modification that we make to the original network is that for every image channel we learn a different RBF mixture. The reason for this is that due to the color transformation the three resulting channels have different SNRs that need to be correctly accounted for. Finally, it is important to note that all the image channels share the same filters of the convolutional and weighted-sum layers and their transposes. The reasoning here is that this way the network can better exploit the channel correlations. A by-product of the specific network design is that the search for similar patches needs to be performed only once compared to the naive implementation that would demand it to be computed independently for each channel. In addition, since this operation is computed only once from the noisy input and then it is re-used in all the network stages, the processing of the color channels can take place in a completely decoupled way and therefore the network admits a very efficient parallel implementation.
{\begin{figure*}[t]
\centering
\begin{tabular}{@{} c @{ } c @{ } c @{ } c @{ } c @{ } c @{ } }
 \begin{overpic}[width=.165\linewidth]{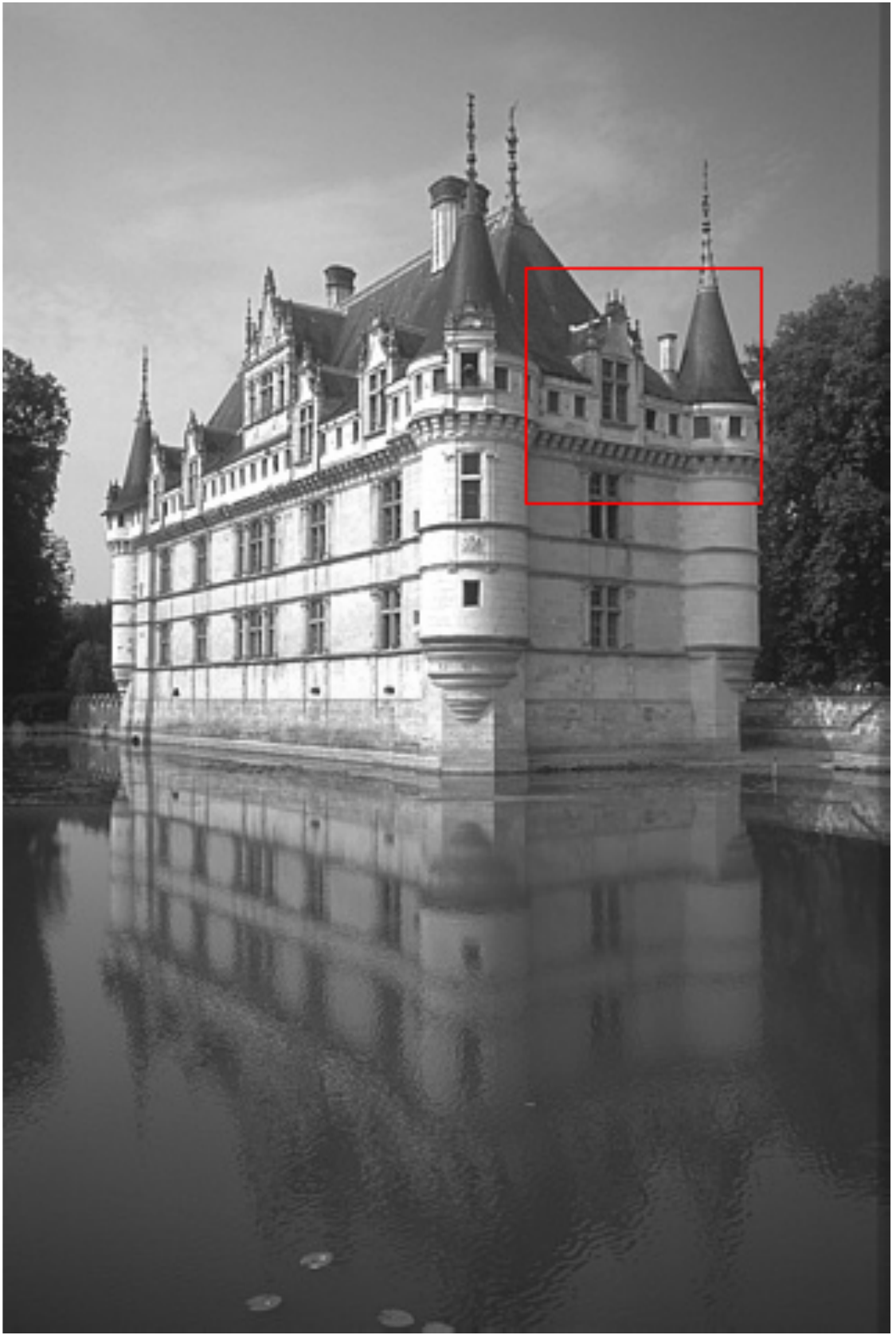}
  \put(33,1){\textcolor{red}{\fboxrule=0.5pt\fboxsep=0pt\fbox{\includegraphics[scale=.45,trim=1 2 2 1,clip=true]{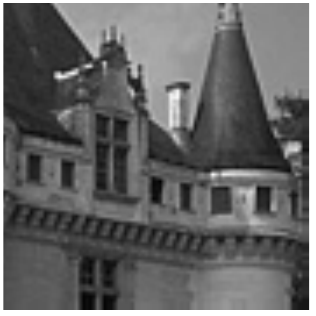}}}}
  \end{overpic}&
 \begin{overpic}[width=.165\linewidth]{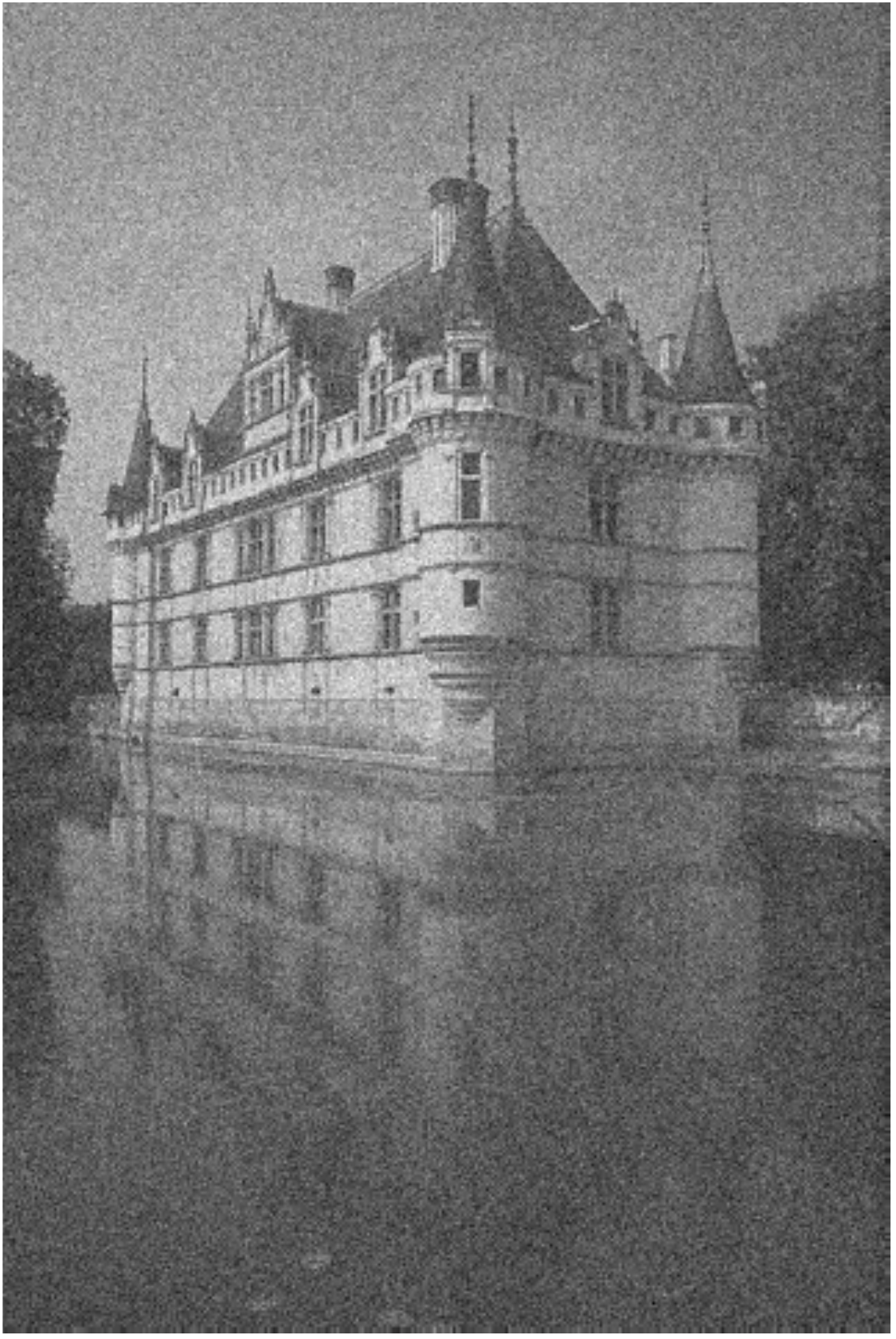}
  \put(33,1){\textcolor{red}{\fboxrule=0.5pt\fboxsep=0pt\fbox{\includegraphics[scale=.45,trim=1 2 2 1,clip=true]{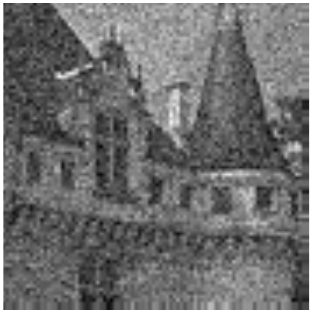}}}}
  \end{overpic}&  
 \begin{overpic}[width=.165\linewidth]{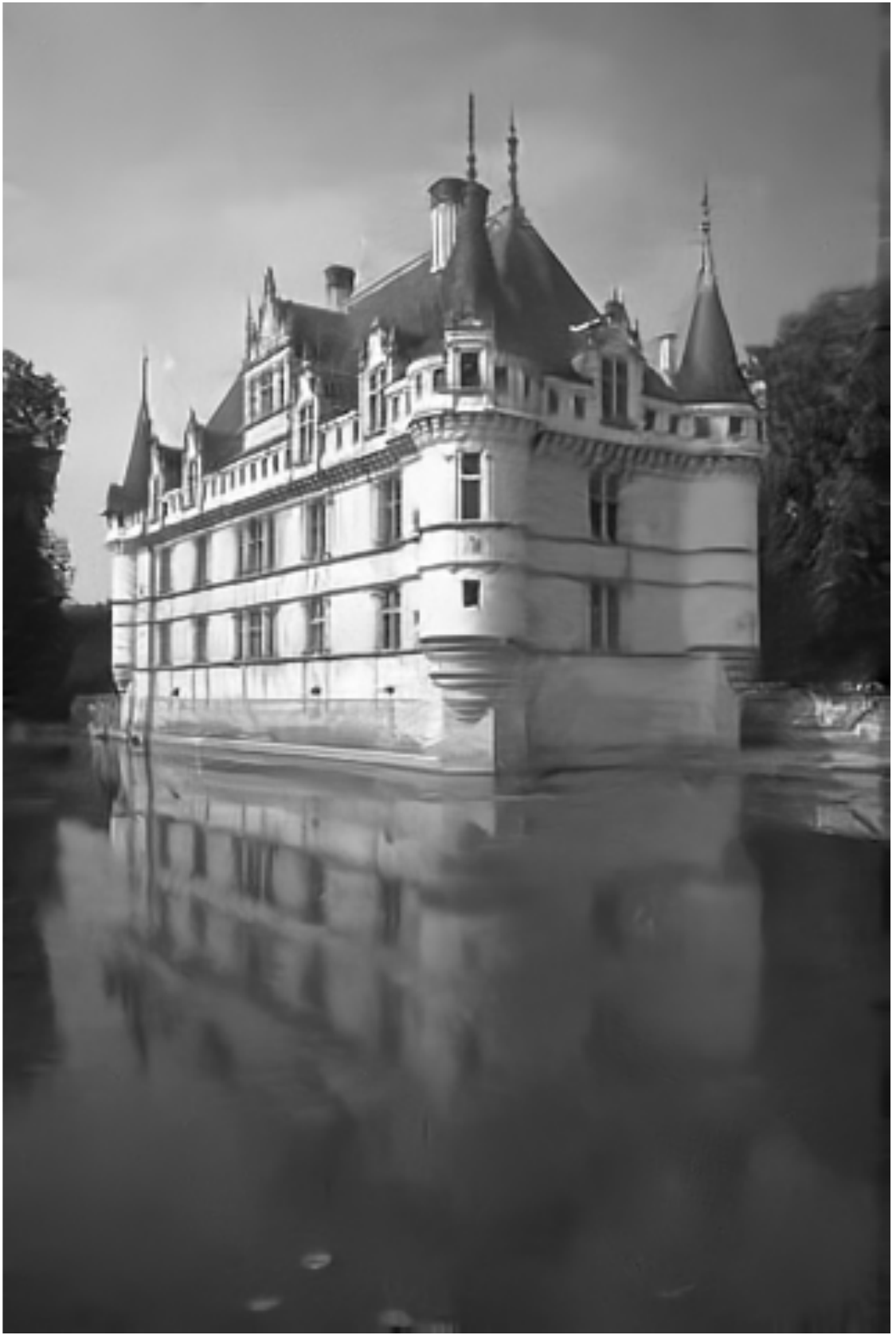}
  \put(33,1){\textcolor{red}{\fboxrule=0.5pt\fboxsep=0pt\fbox{\includegraphics[scale=.45,trim=1 2 2 1,clip=true]{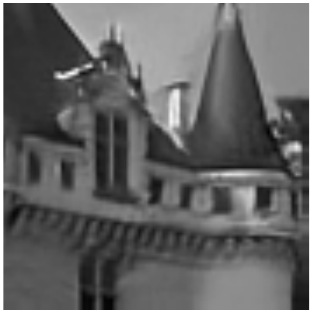}}}}
  \end{overpic}&    
 \begin{overpic}[width=.165\linewidth]{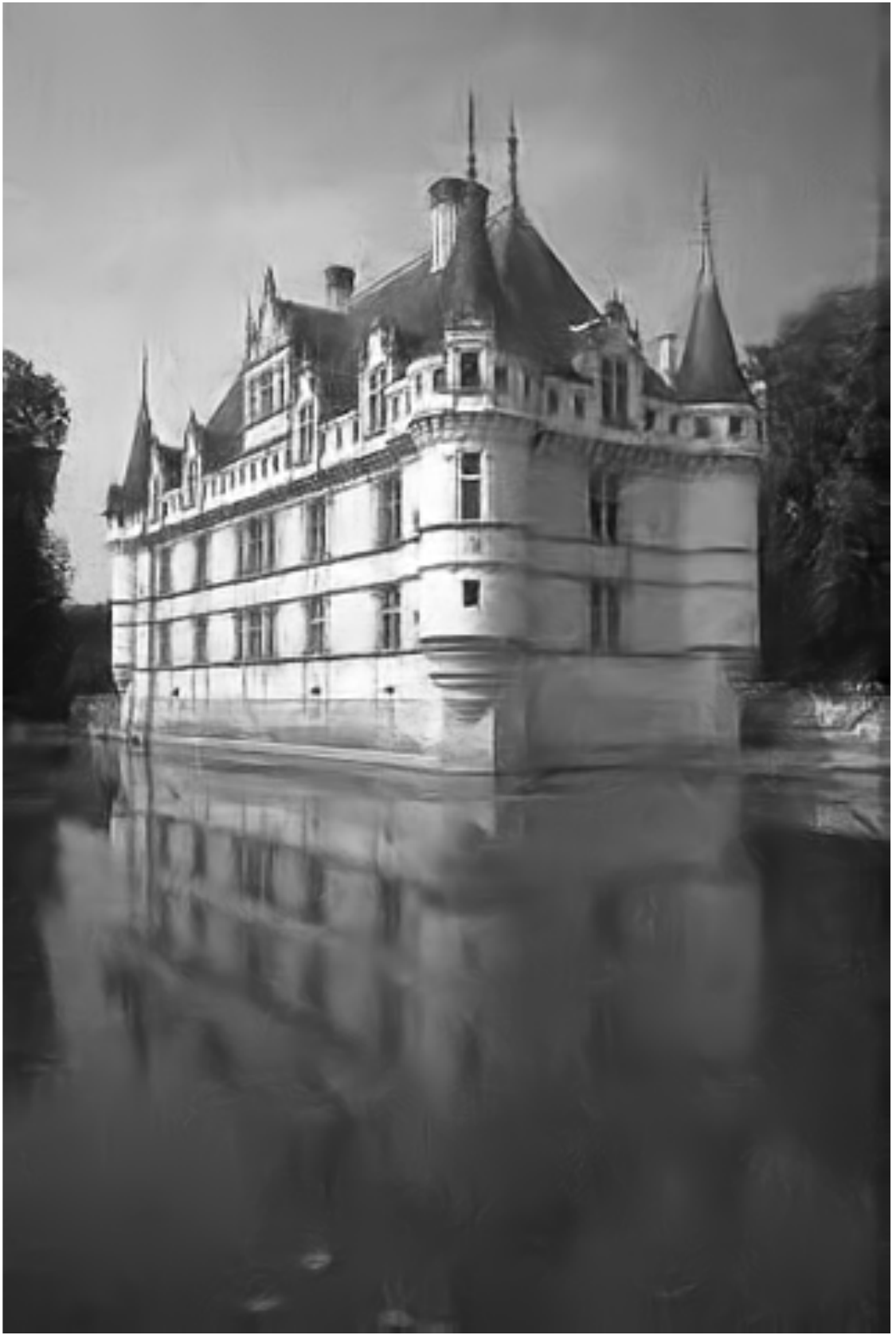}
  \put(33,1){\textcolor{red}{\fboxrule=0.5pt\fboxsep=0pt\fbox{\includegraphics[scale=.45,trim=1 2 2 1,clip=true]{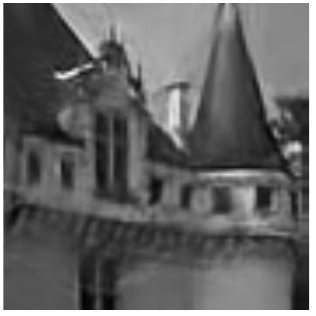}}}}
  \end{overpic}& 
 \begin{overpic}[width=.165\linewidth]{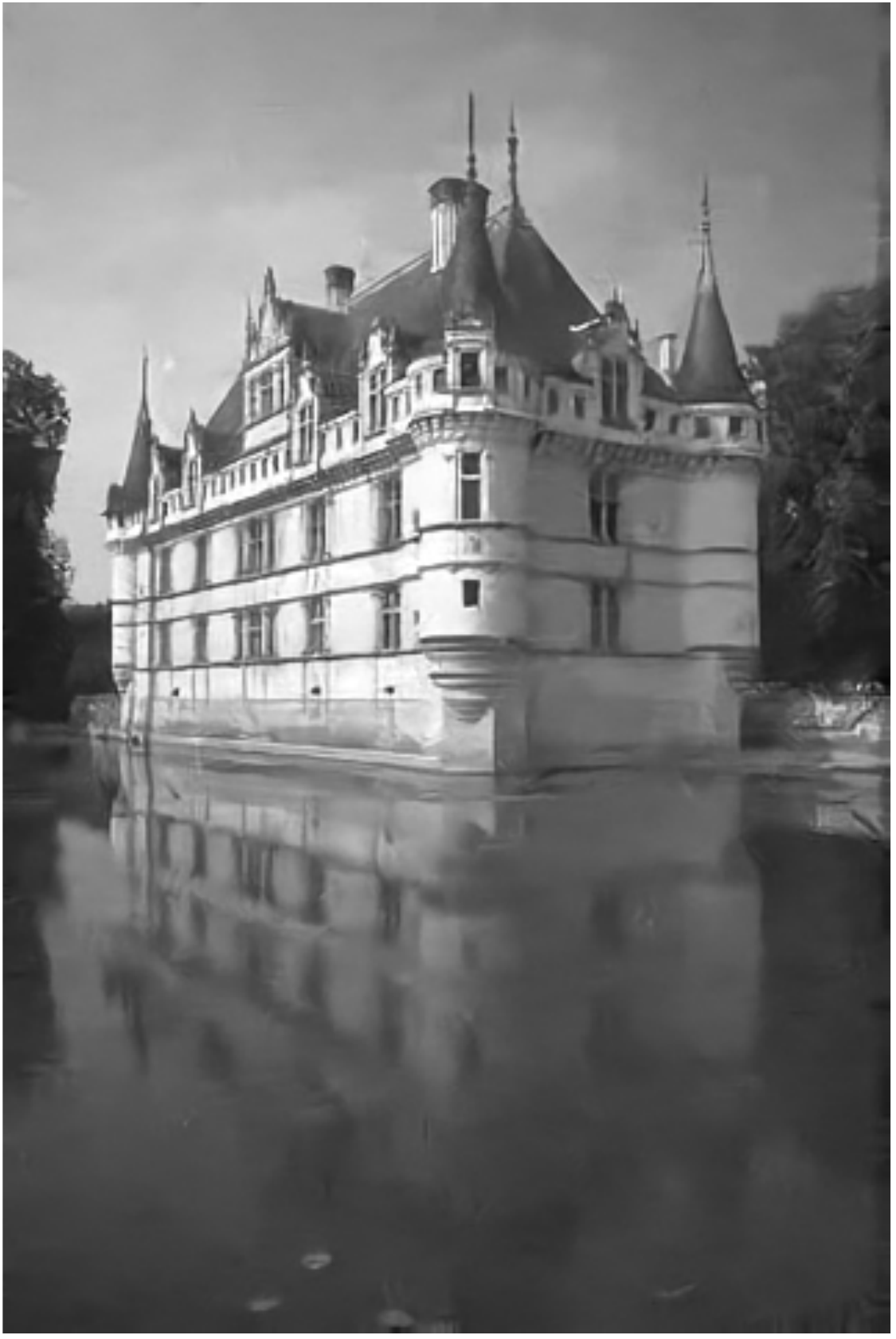}
  \put(33,1){\textcolor{red}{\fboxrule=0.5pt\fboxsep=0pt\fbox{\includegraphics[scale=.45,trim=1 2 2 1,clip=true]{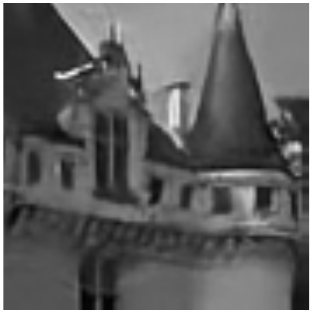}}}}
  \end{overpic}& 
 \begin{overpic}[width=.165\linewidth]{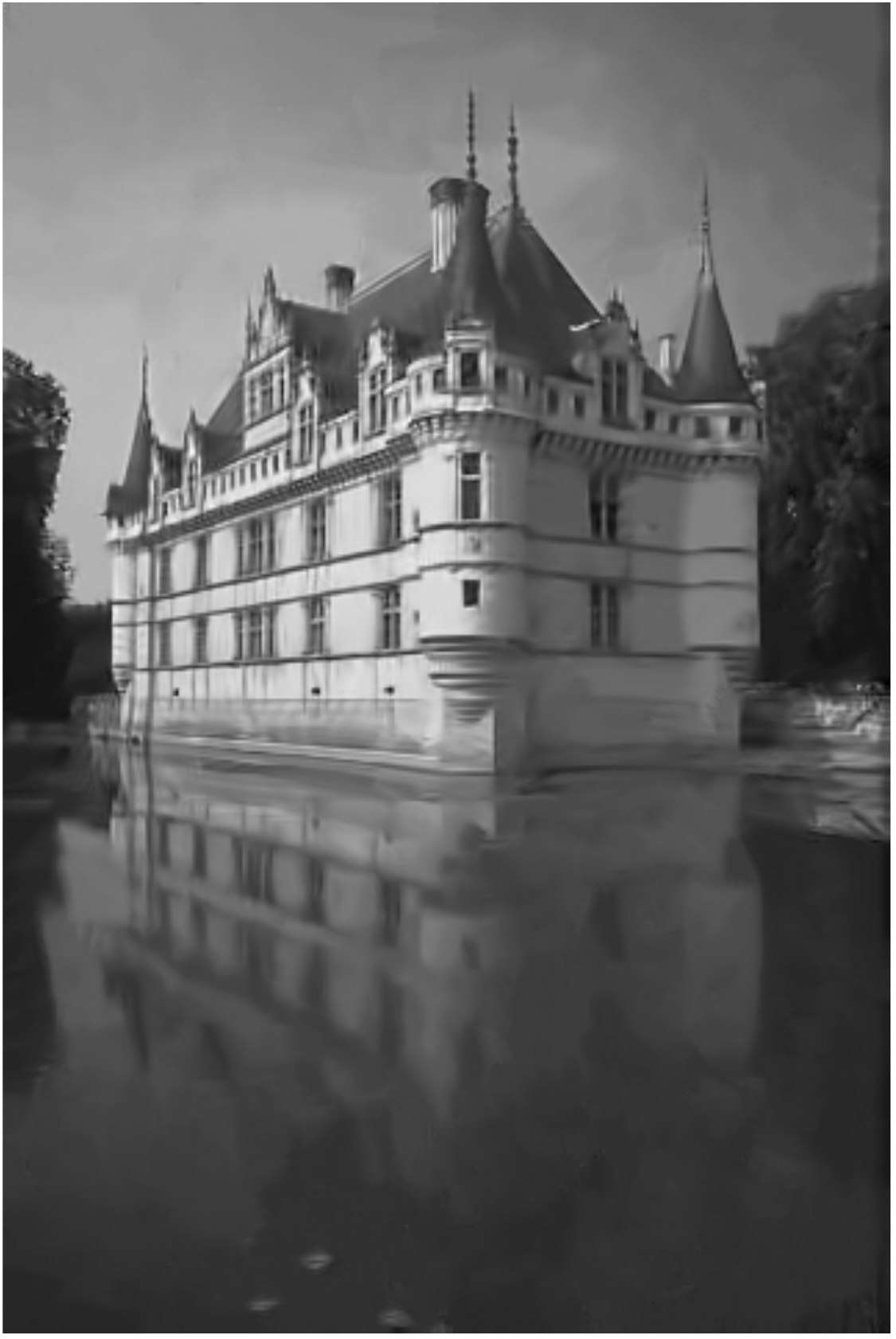}
  \put(33,1){\textcolor{red}{\fboxrule=0.5pt\fboxsep=0pt\fbox{\includegraphics[scale=.45,trim=1 2 2 1,clip=true]{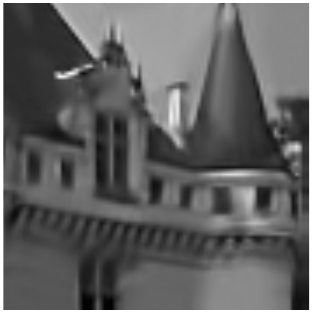}}}}
  \end{overpic}\\     
    (a) & (b) & (c) & (d) & (e) & (f)
\end{tabular}
   \caption{Grayscale image denoising. (a) Original image, (b) Noisy image corrupted with Gaussian noise ($\sigma = 25$) ; $\operatorname{PSNR} = 20.16 \text{ dB}$. (c) Denoised image using $\operatorname{NLNet}_{7\times 7}^5$ ; $\operatorname{PSNR} = 29.95 \text{ dB}$. (d) Denoised image using $\operatorname{TNRD}_{7\times 7}^5$~\cite{Chen2016} ; $\operatorname{PSNR} = 29.72 \text{ dB}$. (e) Denoised image using MLP~\cite{Burger2012} ; $\operatorname{PSNR} = 29.76 \text{ dB}$. (f) Denoised image using WNNM~\cite{Gu2014} ; $\operatorname{PSNR} = 29.76 \text{ dB}$.}
   \label{fig:GrayComp}
\end{figure*}

\begin{figure*}[t]
\centering
\begin{tabular}{@{} c @{ } c @{ } c @{ } c @{ } }
 \begin{overpic}[width=.25\linewidth]{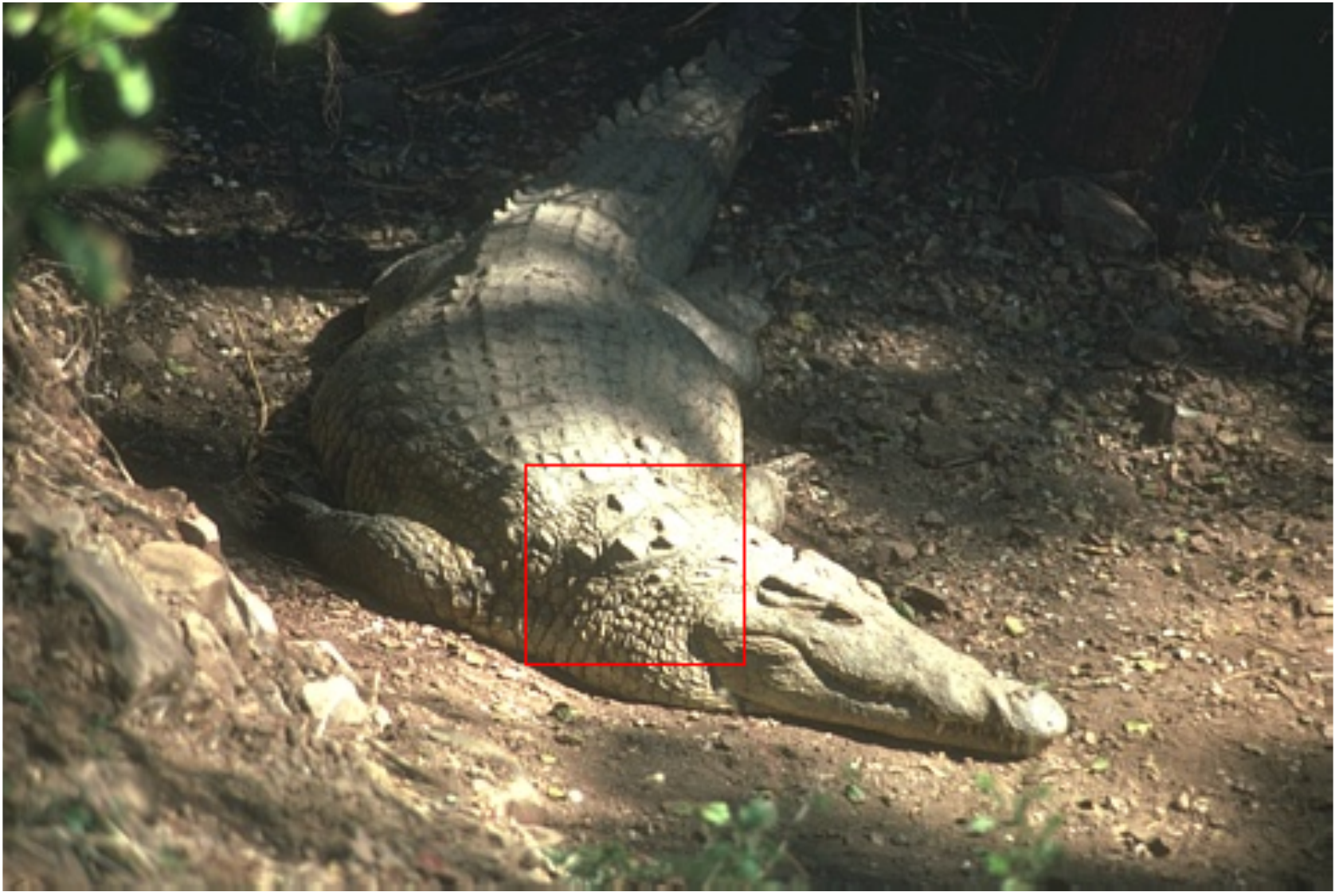}
  \put(65.5,35.5){\textcolor{red}{\fboxrule=0.5pt\fboxsep=0pt\fbox{\includegraphics[scale=.5,trim=1 2 2 1,clip=true]{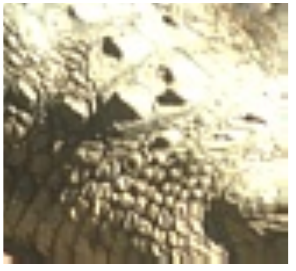}}}}
  \end{overpic}&
 \begin{overpic}[width=.25\linewidth]{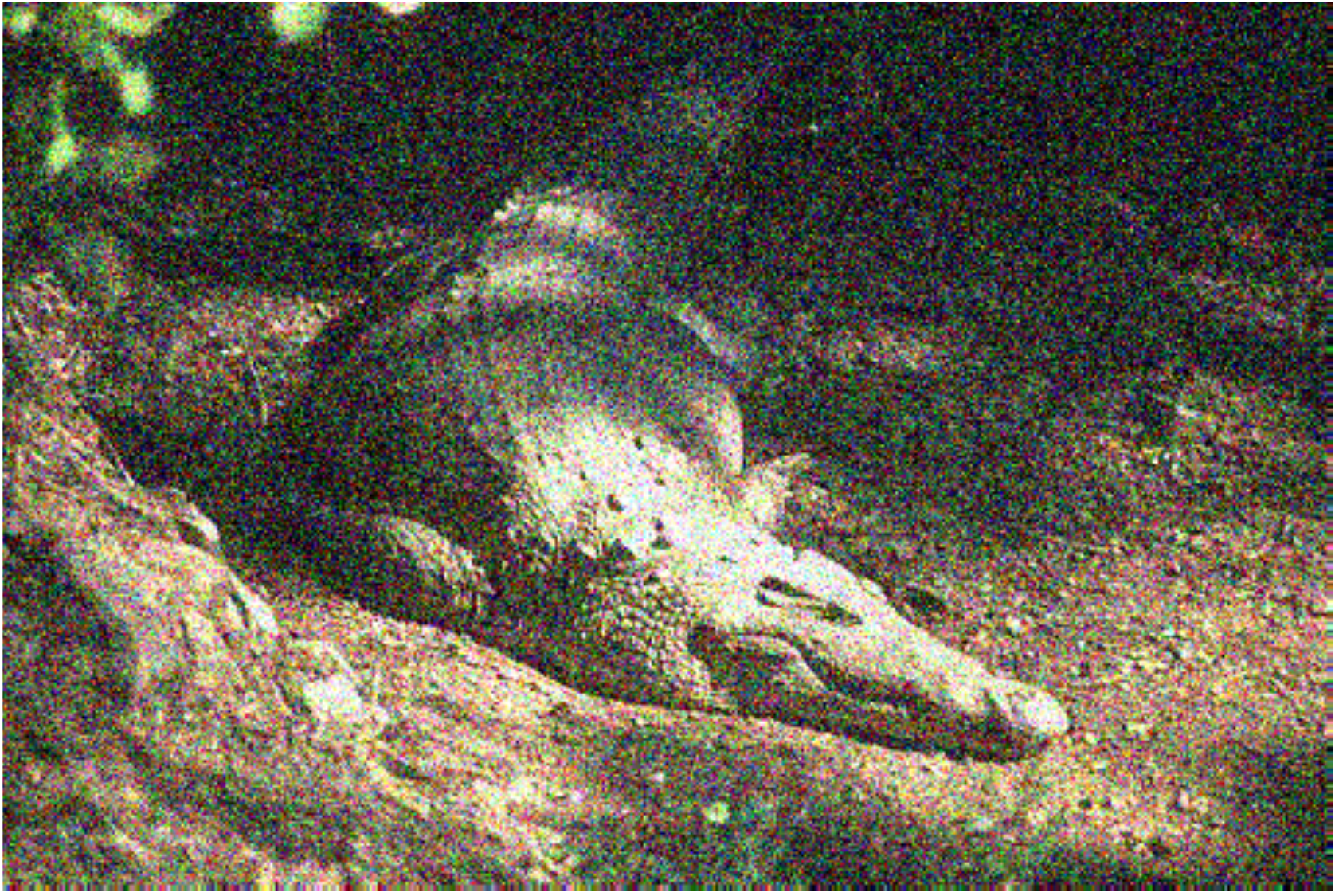}
  \put(65.5,35.5){\textcolor{red}{\fboxrule=0.5pt\fboxsep=0pt\fbox{\includegraphics[scale=.5,trim=1 2 2 1,clip=true]{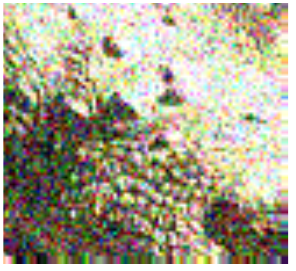}}}}
  \end{overpic}&  
 \begin{overpic}[width=.25\linewidth]{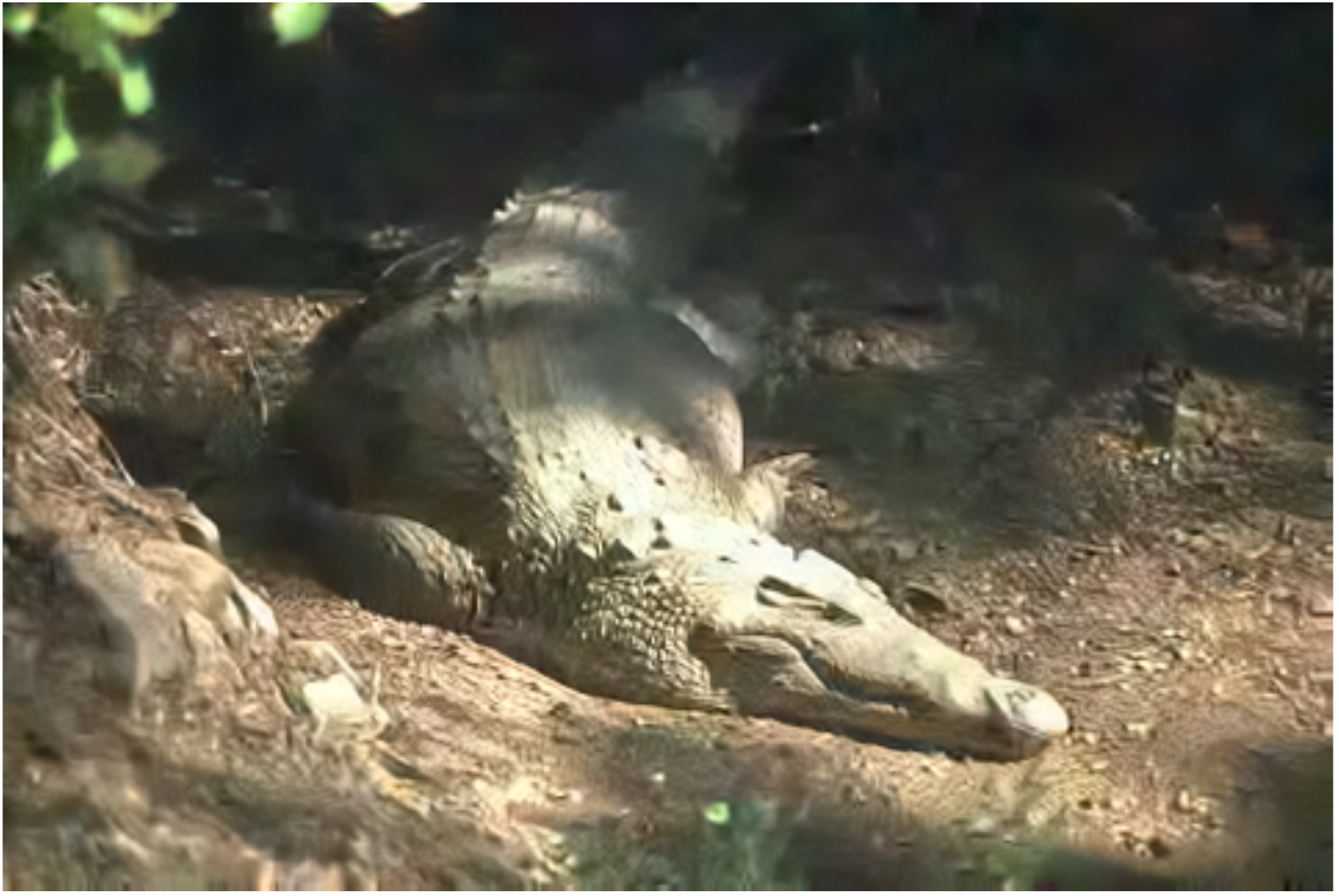}
  \put(65.5,35.5){\textcolor{red}{\fboxrule=0.5pt\fboxsep=0pt\fbox{\includegraphics[scale=.5,trim=1 2 2 1,clip=true]{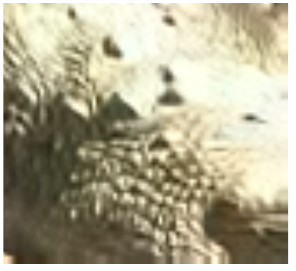}}}}
  \end{overpic}&    
 \begin{overpic}[width=.25\linewidth]{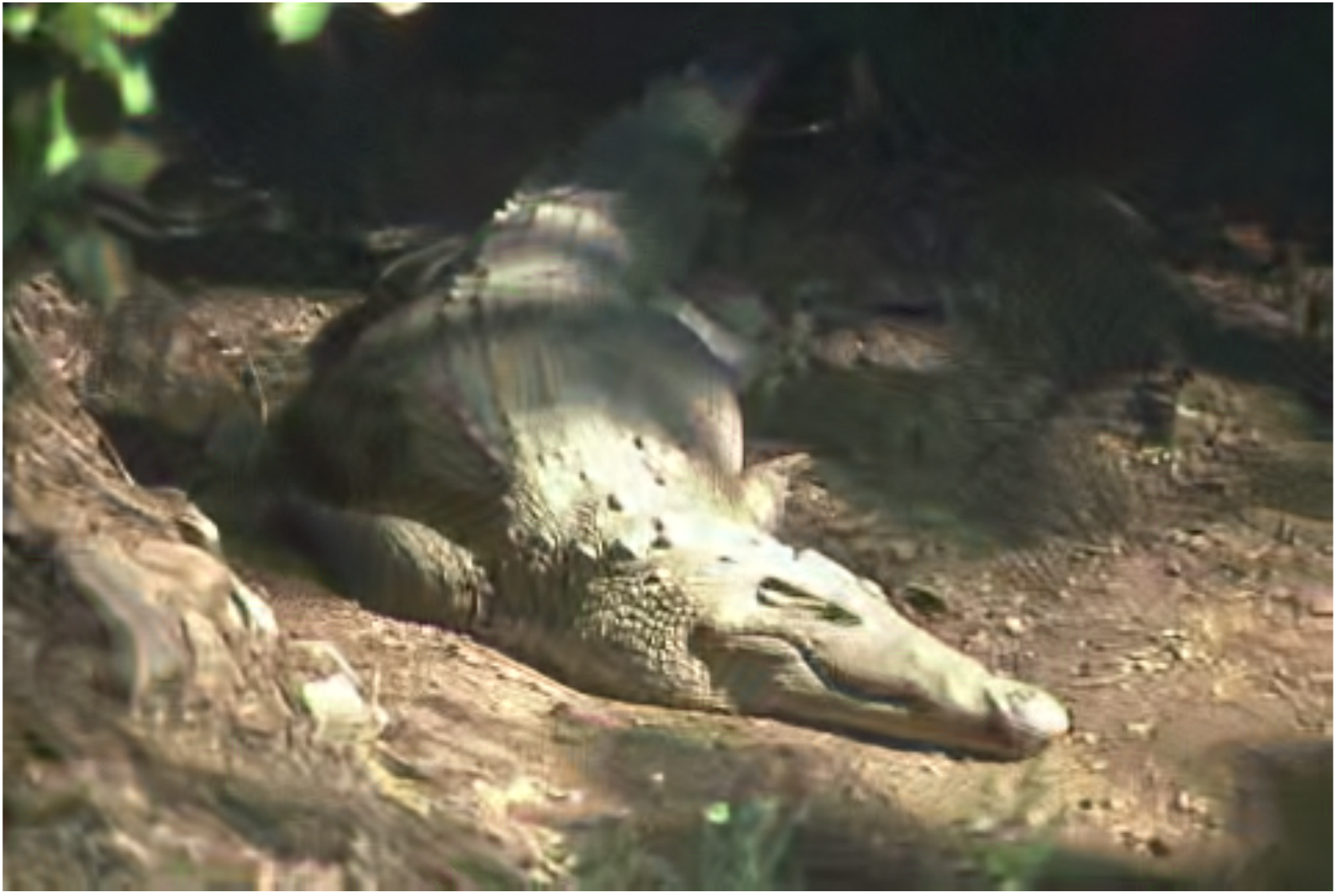}
  \put(65.5,35.5){\textcolor{red}{\fboxrule=0.5pt\fboxsep=0pt\fbox{\includegraphics[scale=.5,trim=1 2 2 1,clip=true]{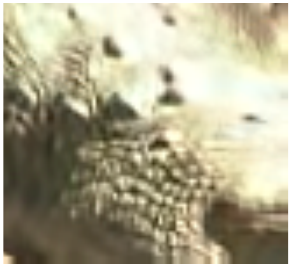}}}}
  \end{overpic}\\     
    (a) & (b) & (c) & (d)
\end{tabular}
   \caption{Color image denoising. (a) Original image, (b) Noisy image corrupted with Gaussian noise ($\sigma = 50$) ; $\operatorname{PSNR} = 14.15 \text{ dB}$. (c) Denoised image using $\operatorname{CNLNet}_{5\times 5}^5$ ; $\operatorname{PSNR} = 26.06 \text{ dB}$. (d) Denoised image using  CBM3D~\cite{Dabov2007} ; $\operatorname{PSNR} = 25.65 \text{ dB}$.}
   \label{fig:ColorComp}
\end{figure*}}

\renewcommand{\tabcolsep}{.09cm}
\begin{table*}[t]
 \begin{tabular}{cccccccccccc}\hline\hline
\multicolumn{2}{c|}{Noise}& \multicolumn{10}{c}{Methods} \\
\multicolumn{2}{c|}{$\sigma$ (std.)}&
\multicolumn{1}{c|}{\small BM3D~\cite{Dabov2007}}&
\multicolumn{1}{c|}{\small LSSC~\cite{Mairal2009}}&
\multicolumn{1}{c|}{\small EPLL~\cite{Zoran2011}}&
\multicolumn{1}{c|}{\small WNNM~\cite{Gu2014}}&
\multicolumn{1}{c|}{\small $\mathrm{CSF}_{7\times 7}^5$~\cite{Schmidt2014}}&
\multicolumn{1}{c|}{\small $\mathrm{TNRD}_{7\times 7}^5$~\cite{Chen2016}} &
\multicolumn{1}{c|}{\small \cellcolor[gray]{0.8}$\mathrm{DGCRF}_8$\cite{Vemulapalli2016}} &
\multicolumn{1}{c|}{\small \cellcolor[gray]{0.8}MLP\cite{Burger2012}}&
\multicolumn{1}{c|}{\small \cellcolor[gray]{0.8}$\mathrm{NLNet}_{5\times 5}^5$}&
\multicolumn{1}{c}{\small \cellcolor[gray]{0.8}$\mathrm{NLNet}_{7\times 7}^5$}\\

\hline

\multicolumn{2}{c|}{\small 15}
&\multicolumn{1}{c|}{\small 31.08}&\multicolumn{1}{c|}{\small 31.27}&\multicolumn{1}{c|}{\small 31.19}
&\multicolumn{1}{c|}{\small 31.37}
&\multicolumn{1}{c|}{\small 31.24}&\multicolumn{1}{c|}{\small 31.42}&\multicolumn{1}{c|}{\small \cellcolor[gray]{0.8}31.43}
&\multicolumn{1}{c|}{\small \cellcolor[gray]{0.8}--}&\multicolumn{1}{c|}{\small \cellcolor[gray]{0.8}\textbf{31.49}}&\multicolumn{1}{c}{\small \cellcolor[gray]{0.8}\textbf{31.52}}\\

\multicolumn{2}{c|}{\small 25}
&\multicolumn{1}{c|}{\small 28.56}&\multicolumn{1}{c|}{\small 28.70}&\multicolumn{1}{c|}{\small 28.68}
&\multicolumn{1}{c|}{\small 28.83}
&\multicolumn{1}{c|}{\small 28.72}&\multicolumn{1}{c|}{\small 28.92}&\multicolumn{1}{c|}{\small \cellcolor[gray]{0.8}28.89}
&\multicolumn{1}{c|}{\small \cellcolor[gray]{0.8}28.96}&\multicolumn{1}{c|}{\small \cellcolor[gray]{0.8}\textbf{28.98}}&\multicolumn{1}{c}{\small \cellcolor[gray]{0.8}\textbf{29.03}}\\

\multicolumn{2}{c|}{\small 50}
&\multicolumn{1}{c|}{\small 25.62}&\multicolumn{1}{c|}{\small 25.72}&\multicolumn{1}{c|}{\small 25.67}
&\multicolumn{1}{c|}{\small 25.83}
&\multicolumn{1}{c|}{\small --}&\multicolumn{1}{c|}{\small 25.96}&\multicolumn{1}{c|}{\small \cellcolor[gray]{0.8}--}
&\multicolumn{1}{c|}{\small \cellcolor[gray]{0.8}\textbf{26.02}}&\multicolumn{1}{c|}{\small \cellcolor[gray]{0.8}25.99}&\multicolumn{1}{c}{\small \cellcolor[gray]{0.8}\textbf{26.07}}\\

\hline\hline
\end{tabular}
\\
\caption{Grayscale image denoising comparisons for three different noise levels over the standard set of 68~\cite{Roth2009} Berkeley images. The restoration performance is measured in terms of average PSNR (in dB) and the best two results are highlighted in bold. The left part of the table is quoted from Chen \etal ~\cite{Chen2016}, while the results of $\operatorname{DGCRF_8}$ are taken from ~\cite{Vemulapalli2016} .}
\label{tab:GrayComp}
\end{table*}

\vspace{-.15cm}
\section{Discriminative Network Training}
\label{sec:NetTrain}
\vspace{-.1cm}
We train our network, which consists of $S$ stages, for grayscale and color image denoising, where the images are corrupted by i.i.d Gaussian noise. The network parameters $\bm{\Theta} = \br{\bm{\Theta}^1, \hdots, \bm{\Theta}^S}$, where $\bm{\Theta}^t = \cbr{\gamma^t, \bm{\pi}^t, \m F^t, \m{W}^t}$ denotes the set of parameters for the $t$-th stage, are learned using a loss-minimization strategy given $Q$ pairs of training data $\cbr{\m y_{\pr{q}}, \m x_{\pr{q}}}_{q=1}^{Q}$, where $\m y_{\pr{q}}$ is a noisy input and $\m x_{\pr{q}}$ is the corresponding ground-truth image. To achieve an increased capacity for the network, we learn different parameters for each stage. Therefore, the overall architecture of the network does not exactly map to the proximal gradient method but rather to an adaptive version. Nevertheless, in each stage the convolution and deconvolution layers share the same filter parameters and, thus, they correspond to proper proximal gradient iterations.

Since the objective function that we need to minimize is non-convex, in order to avoid getting stuck in a bad local-minima but also to speed-up the training, initially we learn the network parameters by following a greedy-training strategy. The same approach has been followed in~\cite{Schmidt2014,Chen2016}. In this case, we minimize the cost 
\bal
\mc{L}\pr{\bm{\Theta}^t} = \suml_{q=1}^Q \ell\pr{\hat{\m{x}}_{\pr{q}}^{t},\m x_{\pr{q}}},
\label{eq:Objective}
\eal
where $\hat{\m{x}}_{\pr{q}}^t$ is the output of the $t$-th stage and the loss function $\ell$ corresponds to the negative peak signal-to-noise-ratio (PSNR). This is computed as
\bal
\ell\pr{\m y,\m x} =  -20\log_{10}\pr{\frac{P_{\operatorname{int}}\sqrt{N}}{\norm{\m y - \m x}{2}}},
\eal
where $N$ is the total number of pixels of the input images and $P_{\operatorname{int}}$ is the maximum intensity level (\ie $P_{\operatorname{int}} = 255$ for grayscale images and $P_{\operatorname{int}} = 1$ for color images).

\renewcommand{\tabcolsep}{.09cm}
\begin{table}
 \begin{tabular}{cccccccc}\hline\hline
\multicolumn{3}{c|}{Noise}& \multicolumn{4}{c}{Methods} \\
\multicolumn{3}{c|}{$\sigma$ (std.)}&
\multicolumn{1}{c|}{\small $\mathrm{TNRD}_{7\times 7}^5$~\cite{Chen2016}} &
\multicolumn{1}{c|}{\small MLP~\cite{Burger2012}}&
\multicolumn{1}{c|}{\small CBM3D~\cite{Dabov2007}}&
\multicolumn{1}{c}{\small $\mathrm{CNLNet}_{5\times 5}^5$}\\
 \hline

\multicolumn{3}{c|}{15}
&\multicolumn{1}{c|}{31.37}&\multicolumn{1}{c|}{--}
&\multicolumn{1}{c|}{33.50}&\multicolumn{1}{c}{\textbf{33.69}}\\

\multicolumn{3}{c|}{25}
&\multicolumn{1}{c|}{28.88}&\multicolumn{1}{c|}{28.92}
&\multicolumn{1}{c|}{30.69}&\multicolumn{1}{c}{\textbf{30.96}}\\

\multicolumn{3}{c|}{50}
&\multicolumn{1}{c|}{25.94}&\multicolumn{1}{c|}{26.00}
&\multicolumn{1}{c|}{27.37}&\multicolumn{1}{c}{\textbf{27.64}}\\

\hline\hline
\end{tabular}
\\
\caption{Color image denoising comparisons for three different noise levels over the standard set of 68~\cite{Roth2009} Berkeley images. The restoration performance is measured in terms of average PSNR (in dB) and the best result is highlighted in bold.} 
\label{tab:ColorComp}
\end{table}

To minimize the objective function in Eq.~\eqref{eq:Objective} w.r.t the parameters $\bm{\Theta}^t$ we employ the L-BFGS algorithm~\cite{Nocedal2006} (we use the available implementation of~\cite{Schmidt2005}). The L-BFGS is a Quasi-Newton method and therefore it requires the gradient of $\mc{L}$ w.r.t $\bm{\Theta}^t$. This can be computed using the chain-rule as
\bal
\frac{\partial \mc{L}\pr{\bm{\Theta}^t}}{\partial \bm{\Theta}^t} = \suml_{q=1}^Q\frac{\partial \hat{\m x}^t_{\pr{q}}}{\partial \bm{\Theta}^t} \cdot \frac{\partial \ell\pr{\hat{\m x}^t_{\pr{q}},\m x_{\pr{q}}}}{\partial \hat{\m x}^t_{\pr{q}}}
\eal
where 
$
\frac{\partial \ell\pr{\m y,\m x}}{\partial \m y} = \frac{20}{\log{10}} \frac{\pr{\m y -  \m x}}{\norm{\m y -\m x}{2}^2},
$
and $\frac{\partial \hat{\m x}^t_{\pr{q}}}{\partial \bm{\Theta}^t}$ is the Jacobian of the output of the $t$-th stage, which can be computed using Eq.~\eqref{eq:proxIter}. We omit the details about the computation of the derivatives w.r.t specific network parameters and we provide their derivations in the appendix. Here, it suffices to say that the gradient of the loss function can be efficiently computed using the back-propagation algorithm~\cite{Rumelhart1986}, which is a clever implementation of the chain-rule. 

For the greedy-training we run 100 L-BFGS iterations to learn the parameters of each stage independently. Then we use the learned parameters as initialization of the network and we train all the stages jointly. The joint training corresponds to minimizing the cost function
\bal
\mc{L}\pr{\bm{\Theta}} = \suml_{q=1}^Q \ell\pr{\hat{\m{x}}_{\pr{q}}^{S},\m x_{\pr{q}}},
\label{eq:JointObjective}
\eal 
w.r.t to all the parameters of the network $\bm{\Theta}$. This cost function does not take into account anymore the intermediate results but only depends on the final output of the network $\hat{\m{x}}_{\pr{q}}^{S}$. In this case we run 400 L-BFGS iterations to refine the result that we have obtained from the greedy-training. Similarly to the previous case, we still employ the back-propagation algorithm to compute the required gradients.

\section{Experiments}
To train our grayscale and color non-local models we generated the training data using the Berkeley segmentation dataset (BSDS)~\cite{Martin2001} which consists of 500 images. We split these images in two sets, a training set which consists of 400 images and the validation/test set which consists of the remaining 100 images. All the images were randomly cropped and their resulting size was $180 \times 180$ pixel.  We note that the 68 BSDS images of~\cite{Roth2009} that are used for the comparisons reported in Tables~\ref{tab:GrayComp} and ~\ref{tab:ColorComp} are strictly excluded from the training set. The proposed models were trained on a NVIDIA Tesla K-40 GPU and the software we used for training and testing was built on top of MatConvnet~\cite{Vedaldi2015}.

\vspace{-.45cm}
\paragraph{Grayscale denosing} Following the strategy described in Section~\ref{sec:NetTrain}, we have trained 5 stages of two different variations of our model, which we will refer to as $\operatorname{NLNet}_{5\times 5}^5$ and $\operatorname{NLNet}_{7\times 7}^5$. The main difference between them is the configuration of the non-local operator. For the first network we considered patches of size $5\times 5$ while for the second one we have considered slightly larger patches of size $7\times 7$. In both cases, the patch stride is one, that is every pixel in the image is considered as the center of a patch. Consequently, the input images at each network stage are padded accordingly, using symmetric boundaries. In addition, a non-redundant patch-transform, which was learned by training, is applied to every image-patch\footnote{Similarly to variational methods, we do not penalize the DC component of the patch-transform. Therefore, the number of the transform-domain coefficients for a patch of size $P$ is equal to $P-1$.} and the group is formed using the $K=8$ closest neighbors. The similar patches are searched on the noisy input of the network in a window of $31 \times 31$ centered around each pixel. The same group indices are then used for all the stages of the network. 

In Table~\ref{tab:GrayComp} we report comparisons of our proposed $\operatorname{NLNet}_{5\times 5}^5$ and $\operatorname{NLNet}_{7\times 7}^5$ models with several recent state-of-the-art denoising methods on the standard evaluation dataset of 68 images~\cite{Roth2009}. From these results we observe that both our non-local models lead to the best overall performance, with the only exception being the case of $\sigma = 50$ where the MLP denoising method~\cite{Burger2012} achieves a slightly better average PSNR compared to that of $\operatorname{NLNet}_{5\times5}^5$. It worths noting that while $\operatorname{NLNet}_{5\times5}^5$ has a lower capacity (it uses approximately half of the parameters) than both $\opname{CSF}^5_{7\times 7}$ and $\opname{TNRD}^5_{7\times 7}$, it still produces better restoration results in all tested cases. This is attributed to the non-local information that exploits, as opposed to $\opname{CSF}^5_{7\times 7}$ and $\opname{TNRD}^5_{7\times 7}$ which are local models. Representative grayscale denoising results that demonstrate visually the restoration quality of the proposed models are shown in Fig.~\ref{fig:GrayComp}.

\vspace{-.45cm}
\paragraph{Color denoising} Given that in the grayscale case the use of $7 \times 7$ patches did not bring any substantial improvements compared to the use of $5 \times 5$ patches, for the color case we have trained a single configuration of our model, considering only color image patches of size $5\times 5$. Besides the standard differences, as they are described in Section~\ref{sec:colorDen}, between the color and the grayscale versions of the $\operatorname{NLNet}_{5\times 5}^5$ model, the rest of the parameters about the size of the patch-group and the search window remain the same.  

An important remark to make here is that most of the denoising methods that were considered previously have been explicitly designed to treat single-channel images, with the most notable exception being the BM3D, for which it indeed exists a color-version (CBM3D)~\cite{Dabov2007}. In practice, this means that if we need to restore color-images then each of these methods should be applied independently on every image channel. In this case however, their denoising performance does not anymore correspond to state-of-the-art. The reason is that due to their single-channel design they fail to capture the existing correlations between the image channels, and this limitation has a direct impact in the final restoration quality.  This fact is also verified by the color denoising comparisons reported in Table~\ref{tab:ColorComp}. From these results we observe that the TNRD and MLP models, which outperform BM3D for single-channel images, fall behind in restoration performance by more than 1.3 dBs. In fact, for low noise levels CBM3D, which currently produces state-of-the-art results, leads to PSNR gains that exceed 2 dBs. Comparing the proposed non-local model with CBM3D, we observe that $\operatorname{CNLNet}_{5\times 5}^5$ manages to provide better restoration results  for all the reported noise levels, with the PSNR gain ranging approximately between 0.2-0.3 dBs. We are not aware of any other color-denoising method that manages to compete with CBM3D on such large set of images. For a visual inspection of the color restoration performance of $\operatorname{CNLNet}_{5\times 5}^5$ we refer to Figs.~\ref{fig:CNLNet} and~\ref{fig:ColorComp}.

\section{Conclusions and Future Work}
In this work we have proposed a novel network architecture for grayscale and color image denoising. The design of the resulting models has been inspired by non-local variational methods and it exploits the non-local self-similarity property of natural images. We believe that non-local modeling coupled with discriminative learning are the key factors of the improved restoration performance that our models achieve compared to several recent state-of-the-art methods. Meanwhile, the proposed models have direct links to convolutional neural networks and therefore can take full advantage of all the latest advances on parallel GPU computing in deep learning.

We are confident that image restoration is just one of the many inverse imaging problems that our non-local networks can successfully handle. We believe that a very interesting research direction is to investigate the necessary modifications on the design of our current non-local models that would allow them to be efficiently applied to other important reconstruction problems. Another very relevant research question is if it is possible to train a single model that can handle all noise levels.

\appendix
\section{Derivative Calculations}
In this section we provide the necessary derivations for the gradients of the loss function of the network w.r.t the parameters $\m \Theta$. 
We note that for all the derivative calculations we use the denominator layout notation\footnote{For the details of this notation we refer to \scriptsize{\url{https://en.wikipedia.org/wiki/Matrix_calculus\#Denominator-layout_notation}}.}. Further, we recall that in order to learn the parameters $\m \Theta = \cbr{\gamma^t, \bm{\pi}^t, \m F^t, \m{W}^t}_{t=1}^S$ of the network, which consists of $S$ stages, we use two different strategies, namely greedy and joint training. During greedy training we learn the parameters $\bm \Theta^t$ of each stage $t$ of the network independently from the parameters of the other stages by minimizing the loss function of Eq.~\eqref{eq:Objective}. On the other hand, in joint training 
the complete set of the network parameters is learned simultaneously by minimizing the loss function given in Eq.~\eqref{eq:JointObjective}.

\subsection{Single-Stage Parameter Learning}
\label{sec:SingleStageLearning}
First we will consider the greedy training scheme. The results computed here will also be useful in the joint estimation scheme. Since the gradient of the overall loss $\mc{L}$ in Eq.~\eqref{eq:Objective} is decomposed as:
\bal
\frac{\partial \mc{L}\pr{\bm{\Theta}^t}}{\partial \bm{\Theta}^t} = \suml_{q=1}^Q \frac{\partial \ell\pr{\hat{\m x}^t_{\pr{q}},\m x_{\pr{q}}}}{\partial \m \Theta^t},
\eal
hereafter we will consider the case of a single training example $\hat{\m x}^t$.  In order to retain the notation simplicity, in the following computations we will also drop the superscript $t$ from all the variables and use it only when it is necessary. 

As we mentioned earlier, to compute the gradients w.r.t  the network parameters we rely on the chain rule and we get
\bal
\frac{\partial \ell\pr{\hat{\m x},\m x}}{\partial \m \Theta}  = \frac{\partial \hat{\m x}}{\partial \bm{\Theta}} \cdot \frac{\partial \ell\pr{\hat{\m x},\m x}}{\partial \hat{\m x}},
\label{eq:ChainRule}
\eal
where 
\bal
\frac{\partial \ell\pr{\hat{\m x},\m x}}{\partial \hat{\m x}} = \frac{20}{\log{10}} \frac{\pr{\hat{\m x}-  \m x}}{\norm{\hat{\m x} -\m x}{2}^2},
\label{eq:gradLoss_x}
\eal
is a vector of size $N\times 1$.
Now we focus on the computation of the Jacobian of the output of the stage, $\hat{\m x}$, w.r.t the stage parameters. Before doing so, we recall that the output, $\hat{\m x}$, of a stage given an input $\m z$, is computed according to the mapping
\bal
\hat{\m x} &= P_{\mc{C}}\pr{\m z\pr{1-\gamma} + \gamma \m y-\suml_{r=1}^R {\m L_r^\transp}\psi\pr{\m L_r \m z}}.
\label{eq:stageMapMod}
\eal
Note that the Eq.~\eqref{eq:stageMapMod} is just a modified version of Eq.~\eqref{eq:proxIter}, where the variable $\alpha$ is absorbed by the function $\psi$.


The Jacobian of $\hat{\m x}$ w.r.t the parameters of the stage, $\m \Theta$, can now be expressed as
\bal
\frac{\partial \hat{\m x}}{\partial \bm{\Theta}}  = \frac{\partial \bm u}{\partial \bm{\Theta}} \frac{\partial P_{\mc{C}}\pr{\bm u}}{\partial \bm u},
\label{eq:GradT}
\eal
where 
\bal
\bm u = \m z\pr{1-\gamma} + \gamma \m y-\suml_{r=1}^R \m L_r^\transp \psi\pr{ \m L_r \m z}.
\label{eq:u}
\eal
Regarding the projection operator $P_{\mc{C}}\pr{\bm u}$, this is applied element-wise to the vector $\bm u$ and it is defined as: 
\bal
P_{\mc{C}}\pr{u} = \begin{cases}
u, & \text{if } a \le u \le b\\
a,  & \text{if } u < a \\
b,  & \text{if } u > b.
\end{cases}
\eal
The derivative of $P_{\mc{C}}\pr{u}$ w.r.t $u$ is computed as:
\bal
\frac{d P_{\mc{C}}\pr{u}}{d u} = \begin{cases}
1, & \text{if } a \le u \le b\\
0,  & \text{elsewhere},
\end{cases}
\eal
and therefore the Jacobian $\frac{\partial P_{\mc{C}}\pr{\bm u}}{\partial \bm u}$ corresponds to a binary diagonal matrix of size $N \times N$, whose diagonal elements are non-zero only if the corresponding values of $\bm u$ are in the range $\bbmtx a\,\, b \ebmtx$. Now, let us denote as $\bm e$ the $N \times 1$ vector obtained by the matrix vector product of the Jacobian $\frac{\partial P_{\mc{C}}\pr{\bm u}}{\partial \bm u}$ with the gradient $\frac{\partial \ell\pr{\hat{\m x},\m x}}{\partial \hat{\m x}}$, that is

\bal
\bm{e} = \frac{\partial P_{\mc{C}}\pr{\bm u}}{\partial \bm u}\cdot\frac{\partial \ell\pr{\hat{\m x},\m x}}{\partial \hat{\m x}}.
\label{eq:e}
\eal
\paragraph{Weight parameter $\gamma$ :} Using Eq.~\eqref{eq:u} it is straightforward to show that
\bal
\frac{\partial \bm u}{\partial \gamma} = \pr{\m y - \m z}^\transp
\eal
and thus $\frac{\partial \ell\pr{\hat{\m x}, \m x}}{\partial \gamma}$ is computed as
\bal
\frac{\partial \ell\pr{\hat{\m x}, \m x}}{\partial \gamma} = \pr{\m y - \m z}^\transp \cdot \bm e.
\eal

\paragraph{Expansion coefficients $\bm{\pi}$ :} To compute the gradient of the loss function $\ell$ w.r.t to the expansion coefficients $\bm \pi$ of the mixture of Gaussian RBFs, we first express the output of the RBF mixture as a vector inner product. Specifically, it holds that
\bal
\psi_i\pr{x} = \suml_{j=1}^M \pi_{ij}\rho_j\pr{\abs{x-\mu_j}} = \bm{\rho}^\transp\pr{x}\bm{\pi}_i,
\label{eq:RBF}
\eal
where $\bm{\rho}\pr{x}=\bbmtx \rho\pr{\abs{x-\mu_1}}&  \hdots & \rho\pr{\abs{x-\mu_M}}\ebmtx^\transp\in\R^M$. We note that in the definition of $\bm{\rho}\pr{x}$ we have dropped the subscript $j$ from the Gaussian RBF $\rho_j\pr{x} =\exp\pr{-\varepsilon_j x^2}$, since we use a common precision parameter for all the mixture components, \ie $\varepsilon = \varepsilon_j,\, \forall j$.
Based on this notation we can further express $\psi\pr{\m x} = \bbmtx \psi_1\pr{\m{x}_1} & \psi_2\pr{\m{x}_2}& \hdots & \psi_F\pr{\m{x}_F}\ebmtx^{\transp}$ as
\bal
\psi\pr{\m x} = \bm{R}^\transp\pr{\m x} \bm{\pi} 
\label{eq:MultiVarRBF}
\eal
where 
$\bm{\pi} = \bbmtx \bm{\pi}_1^\transp & \hdots & \bm{\pi}_F^\transp \ebmtx^\transp$, $\m x\in R^F$  and 
\bal
\bm{R}^\transp\pr{\m x}= \bbmtx  \bm{\rho}^\transp\pr{\m x_1} & \m{0} & \hdots & \m{0}\\
\m{0} &\bm{\rho}^\transp\pr{\m x_2} &  & \m{0} \\
\vdots & & \ddots & \\
\m{0} & \hdots& \m{0}& \bm{\rho}^\transp\pr{\m x_F}\ebmtx\in\R^{F\times\pr{M\cdot F}}.
\eal
Now, using Eqs.~\eqref{eq:u}, \eqref{eq:RBF} and \eqref{eq:MultiVarRBF} we have
\bal
\bm{u} = \m z\pr{1-\gamma} + \gamma \m y-\suml_{r=1}^R\m{L}_r^{\transp}\bm{R}^\transp\pr{\m L_r \m z}\bm{\pi}
\eal
which directly leads us to compute the Jacobian $\frac{\partial \bm u}{\partial \bm \pi}$ as
\bal
\frac{\partial \bm{u}}{\partial \bm{\pi}} = -\suml_{r=1}^R\bm{R}\pr{\m L_r \m z}{\m{L}_r}.
\label{eq:piDer}
\eal
Finally, combining Eqs.~\eqref{eq:e} and \eqref{eq:piDer} we get

\bal
\frac{\partial \ell\pr{\hat{\m x}, \m x}}{\partial \bm{\pi}} = -\suml_{r=1}^R\bm{R}^\transp\pr{\m L_r \m z}{\m{L}_r}\bm{e}.
\eal

\paragraph{Weighted sum coefficients $\m W$ :} To simplify the computation of the gradient of the loss function w.r.t $\m W$, first we obtain an equivalent expression for the non-local operator $\m L_r$ defined in Eq.~\eqref{eq:NLOperator}. Indeed, the non-local operator can be re-written as
\bal
\m L_r= \suml_{k=1}^K w_k \m F \m P_{i_{r,k}}= \suml_{k=1}^K{w_k \m T_{i_{r,k}}}.
\eal
Plugging the new expression of $\m L_r$ into Eq.~\eqref{eq:u} we get 
\bal
\bm u &= \m z\pr{1-\gamma} + \gamma \m y-\suml_{r=1}^R \suml_{k=1}^K w_k\m T_{i_{r,k}}^\transp
\psi\pr{\suml_{k=1}^Kw_k\m z_{i_{r,k}}},
\label{eq:u_W}
\eal
where $\m z_{i_{r,k}} = \m T_{i_{r,k}}\m z$.
Now, it is straightforward to compute the partial derivative  of $\bm u$ w.r.t each $w_i$. Based on Eq.~\eqref{eq:u_W}, we obtain
\bal
\frac{\partial \bm{u}}{\partial w_i} & = -\suml_{r=1}^R \frac{\partial}{\partial w_i}\pr{ \pr{w_i \m T_{i_{r,i}}^\transp + \suml_{k\ne i}w_k\m T_{i_{r,k}}^\transp}
\psi\pr{\m z_{i_r}}}\nonumber \\
& = -\suml_{r=1}^R\pr{\psi^\transp\pr{\m z_{i_r}}\m T_{i_{r,i}}+\suml_{k=1}^K w_k \m z_{i_{r,i}}^\transp\frac{\partial \psi\pr{\m z_{i_r}}}{\partial \m z_{i_r}}\m T_{i_{r,k}}},
\label{eq:u_wDer}
\eal
where $\m z_{i_r}=\suml_{k=1}^K w_k \m z_{i_{r,k}}$. Note that due to the decoupled formulation of $\psi$, the Jacobian $\frac{\partial \psi\pr{\m z_{i_r}}}{\partial \m z_{i_r}}$ is a diagonal matrix of the form:
\bal
\frac{\partial \psi\pr{\m x}}{\partial \m x} = \bbmtx \frac{\partial\psi_1\pr{\m x_1}}{\partial \m x_1} & 0 & \hdots & 0 \\ 
0& \frac{\partial\psi_2\pr{\m x_2}}{\partial \m x_2} &  & 0 \\
\vdots & & \ddots & \\
0 &  \hdots & 0 & \frac{\partial\psi_F\pr{\m x_F}}{\partial \m x_F} \ebmtx,
\eal
where 
\bal
\frac{\partial \psi_i\pr{x}}{\partial x} = -2\varepsilon\suml_{j=1}^M\pi_{ij}\pr{x-\mu_j}\exp\pr{-\varepsilon\pr{x-\mu_j}^2}.
\eal
Combining Eqs~\eqref{eq:e} and \eqref{eq:u_wDer} we obtain:
\bal
\frac{\partial \ell\pr{\hat{\m x}, \m x}}{\partial w_i} =  -\suml_{r=1}^R\pr{\psi^\transp\pr{\m z_{i_r}}\m T_{i_{r,i}}+\suml_{k=1}^K w_k \m z_{i_{r,i}}^\transp\frac{\partial \psi\pr{\m z_{i_r}}}{\partial \m z_{i_r}}\m T_{i_{r,k}}}\bm{e}.
\eal

%
%

\paragraph{Patch-transform coefficients $\m F$ :} Let us express the matrix $\m F\in\R^{F\times P}$ in terms of its column vectors, \ie 
$\m{F} = \bbmtx \m f_1 & \hdots & \m f_F \ebmtx^\transp$ with $\m f_i\in\R^P\,\forall\, i=1,\hdots,F$. Now, let us also re-write $\m L_r\m z$ as 
\bal
\m L_r \m z & = \pr{\suml_{k=1}^K w_k \m F\,\m P_{i_{r,k}}}\m z \nonumber\\
& = \m F \pr{\suml_{k=1}^K w_k \m P_{i_{r,k}}}\m z = \m F\pr{\m B_r \m z}\nonumber\\
& =  \m F \tilde{\m z}_r = 
\bbmtx \m f_1^\transp \tilde{\m z}_r\\
\vdots\\
\m f_F^\transp \tilde{\m z}_r
\ebmtx.
\label{eq:Lrz}
\eal
Next, we use Eq.~\eqref{eq:Lrz} to re-write Eq.~\eqref{eq:u} as 
\bal
 \bm u &= \m z\pr{1-\gamma} + \gamma \m y-\suml_{r=1}^R \m B_r^\transp \bbmtx \m f_1 & \hdots&  \m f_F\ebmtx \psi\pr{\bbmtx \m f_1^\transp \tilde{\m z}_r\\
\vdots\\
\m f_F^\transp \tilde{\m z}_r
\ebmtx}\nonumber\\
& =
\m z\pr{1-\gamma} + \gamma \m y-\suml_{r=1}^R \m B_r^\transp \bbmtx \m f_1 & \hdots&  \m f_F\ebmtx 
\bbmtx \psi_1\pr{\m f_1^\transp \tilde{\m z}_r}\\
\vdots\\
\psi_F\pr{\m f_F^\transp \tilde{\m z}_r}
\ebmtx\nonumber\\
& = \m z\pr{1-\gamma} + \gamma \m y-\suml_{r=1}^R \m B_r^\transp\pr{\suml_{j=1}^F \m f_j \psi_j\pr{
\m f_j^\transp\tilde{\m z}_r}}.
\eal
This last reformulation of $\bm u$ greatly facilitates the computation of its Jacobian w.r.t $\m f_i, \,\forall i=1,\hdots,F$. Now, we can show that 
\bal
\frac{\partial \bm{u}}{\partial \m{f}_i} = -\suml_{r=1}^R \pr{\m{I}_P \cdot\psi_i\pr{\m f_i\tilde{\m z}_r}+\m f_i\tilde{\m z}_r^\transp\cdot\frac{\partial \psi_i\pr{\m f_i\tilde{\m z}_r}}{\pr{\m f_i\tilde{\m z}_r}}}\m B_r,
\eal
where $\m I_P\in\R^{P\times P}$ is the identity matrix. Consequently, it holds 
\bal
\frac{\partial \ell\pr{\hat{\m x}, \m x}}{\partial \m f_i} = -\suml_{r=1}^R \pr{\m{I}_P \cdot\psi_i\pr{\m f_i\tilde{\m z}_r}+\m f_i\tilde{\m z}_r^\transp\cdot\frac{\partial \psi_i\pr{\m f_i\tilde{\m z}_r}}{\pr{\m f_i\tilde{\m z}_r}}}\m B_r\bm{e}.
\eal

\subsection{Joint Parameter Learning}
In the joint-training scheme the parameters of all the stages of the network are learned simultaneously by minimizing the loss function of Eq.~\eqref{eq:JointObjective} which depends only on the final output of the network $\hat{\m x}^S$. In this case we need to compute the gradient of the loss function $\ell\pr{\hat{\m x}^S,\m x}$ w.r.t the parameters $\bm{\Theta}^t$ of each stage $t$. Using the chain-rule this can be computed as 
\bal
\frac{\partial \ell\pr{\hat{\m x}^S,\m x}}{\partial \bm{\Theta}^t}=\frac{\partial \hat{\m x}^t}{\partial \bm{\Theta}^t}\cdot
\frac{\partial \hat{\m x}^S}{\partial \hat{\m x}^t}\cdot
\frac{\partial \ell\pr{\hat{\m x}^S,\m x}}{\partial \hat{\m x}^S},
\eal
where $\frac{\partial \hat{\m x}^t}{\partial \bm{\Theta}^t}$ is calculated by combining Eq.~\eqref{eq:GradT} and the results of Section~\ref{sec:SingleStageLearning}, while $\frac{\partial \ell\pr{\hat{\m x}^S,\m x}}{\partial \hat{\m x}^S}$ is given by Eq.~\eqref{eq:gradLoss_x}. Therefore, the only remaining Jacobian that we need to compute is $\frac{\partial \hat{\m x}^S}{\partial \hat{\m x}^t}$. This quantity can be computed recursively as 
\bal
\frac{\partial \hat{\m x}^S}{\partial \hat{\m x}^t} = \frac{\partial \hat{\m x}^{t+1}}{\partial \hat{\m x}^t}\cdot\frac{\partial \hat{\m x}^{t+2}}{\partial \hat{\m x}^{t+1}}\cdot\cdot\cdot\frac{\partial \hat{\m x}^S}{\partial \hat{\m x}^{S-1}}.
\eal
Consequently, it suffices to derive the expression for the Jacobian $\frac{\partial \hat{\m x}^{t+1}}{\partial \hat{\m x}^t}$ where $\hat{\m x}^{t+1}$ is obtained from  $\hat{\m x}^{t}$ according to 
\bal
\hat{\m x}^{t+1} &= P_{\mc{C}}\Bigg(\hat{\m x}^t\pr{1-\gamma^{t+1}} + \gamma^{t+1} \m y\nonumber\\
&\quad -\suml_{r=1}^R {\pr{\m L_r^{t+1}}^\transp}\psi^{t+1}\pr{\m L_r ^{t+1}\hat{\m x}^t}\Bigg)\nonumber\\
& = P_{\mc{C}}\pr{\bm u^{t+1}}.
\label{eq:stageMapx}
\eal
Using Eq.~\eqref{eq:stageMapx}, we finally get
\bal
\frac{\partial \hat{\m x}^{t+1}}{\partial \hat{\m x}^t} &= \frac{\partial {\bm u}^{t+1}}{\partial \hat{\m x}^t}\cdot \frac{\partial P_{\mc{C}}\pr{{\bm u}^{t+1}}}{\partial {\bm u}^{t+1}}\nonumber\\
& = \Bigg(\m I_N \pr{1-\gamma^{t+1}}\nonumber\\
&-\suml_{r=1}^R \pr{\m L_r^{t+1}}^\transp\frac{\partial \psi^{t+1}\pr{\hat{\m x}_r^t}}{\partial \hat{\m x}_r^t}\m L_r^{t+1}\Bigg)\m P^{t+1},
\eal
where $\m I_N\in R^{N\times N}$ is the identity matrix, $\hat{\m x}_r^t =\m L_r^{t+1} \hat{\m x}^t$ and $\m P^{t+1} = \frac{P_{\mc{C}}\pr{{\bm u}^{t+1}}}{\partial {\bm u}^{t+1}}.$

\section{Grayscale and Color Image Denoising Comparisons}
In this section we provide additional grayscale and color image denoising results for different noise levels. For grayscale image denoising we compare the performance of our non-local models with TNRD~\cite{Chen2016}, MLP~\cite{Burger2012}, EPLL~\cite{Zoran2011} and BM3D~\cite{Dabov2007}, while for color image denoising we compare our non-local CNN with the state-of-the-art CBM3D method~\cite{Dabov2007}.  Besides the visual comparisons, in the captions of the figures we provide the PSNR score (in dB) of each method to also allow a quantitative comparison.

\begin{figure*}[!t]
\centering
\begin{tabular}{@{} c @{ } c @{ } }
 \includegraphics[width=.5\linewidth]{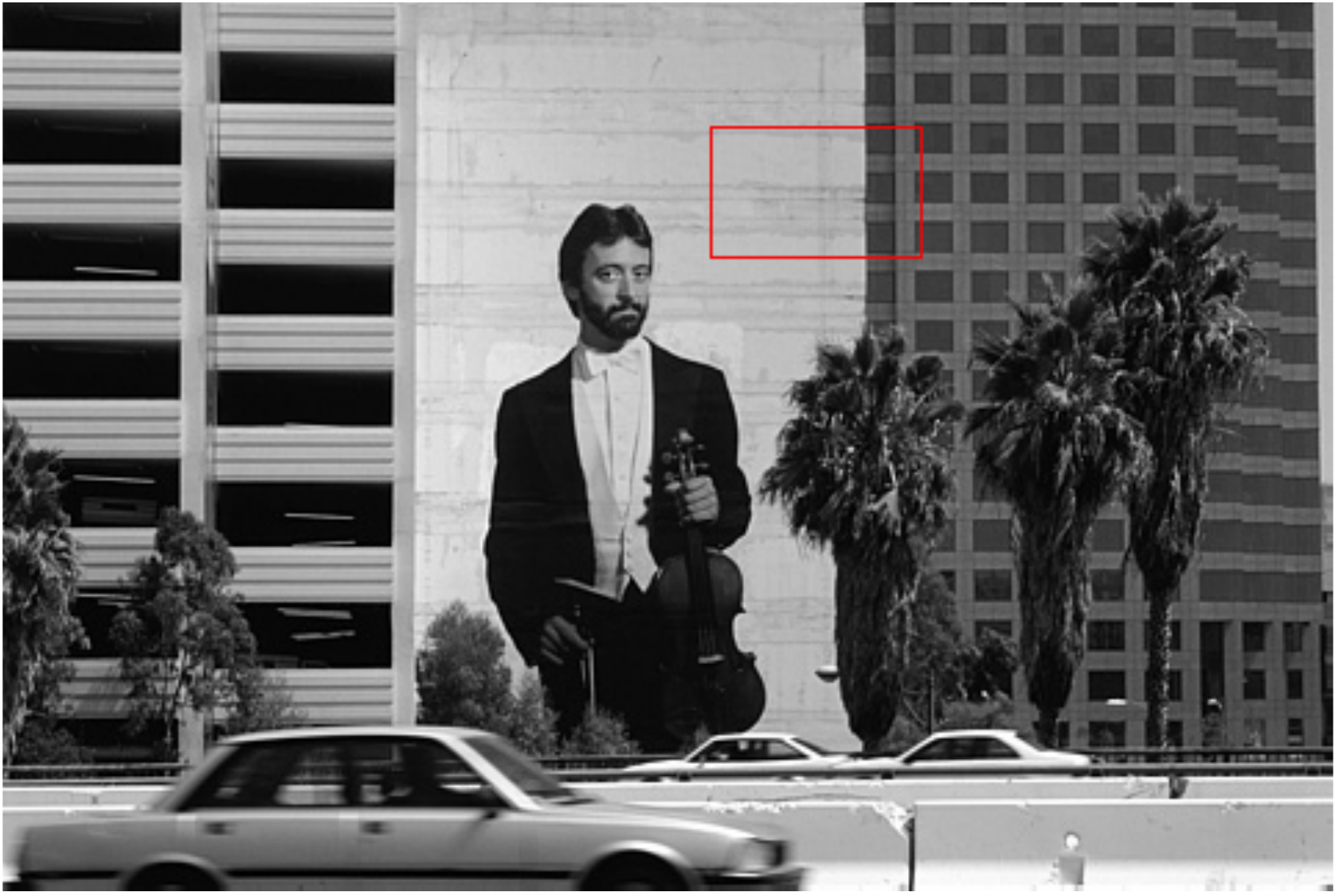}&
\includegraphics[width=.5\linewidth]{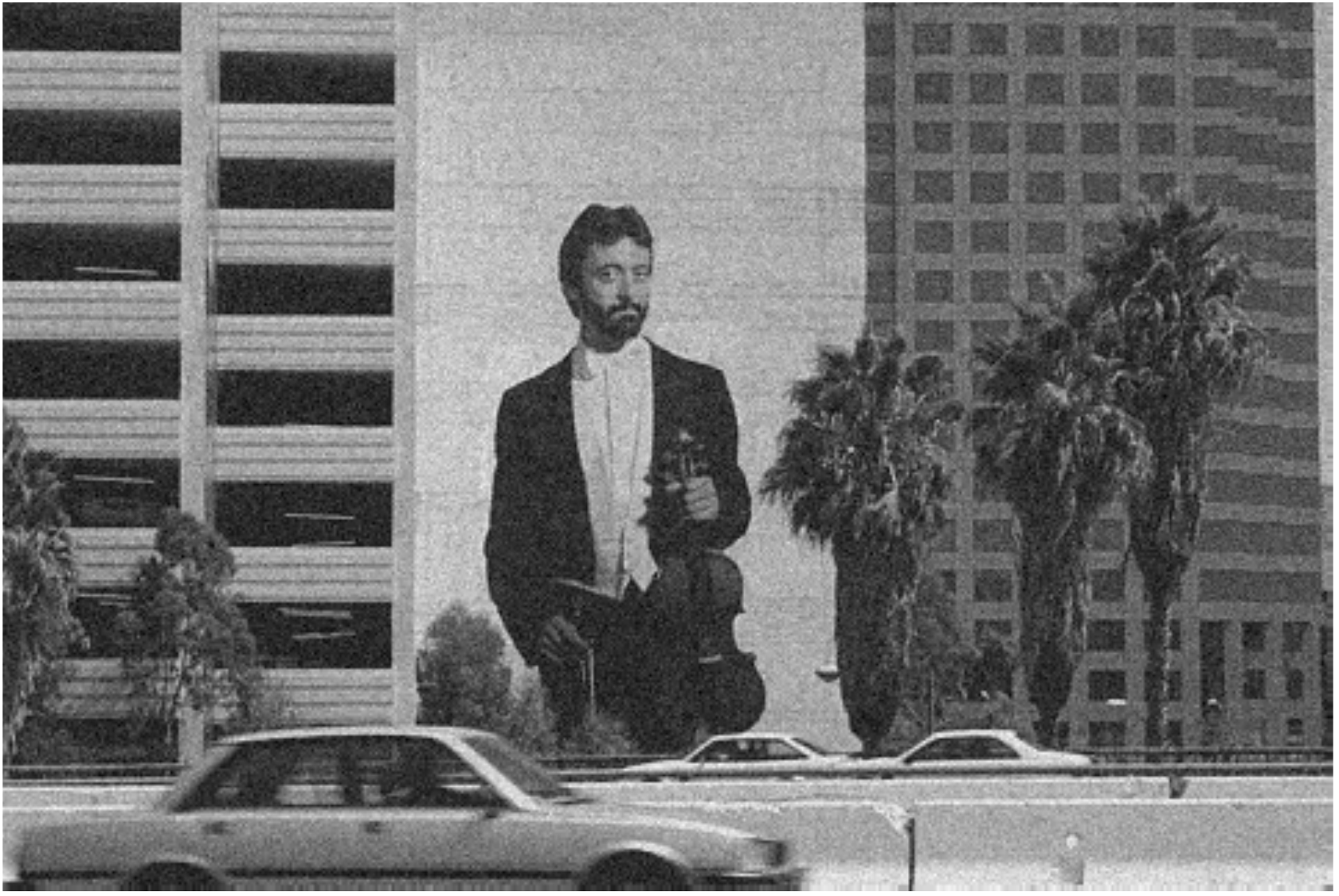}\\
(a) & (b) \\
\includegraphics[width=.5\linewidth]{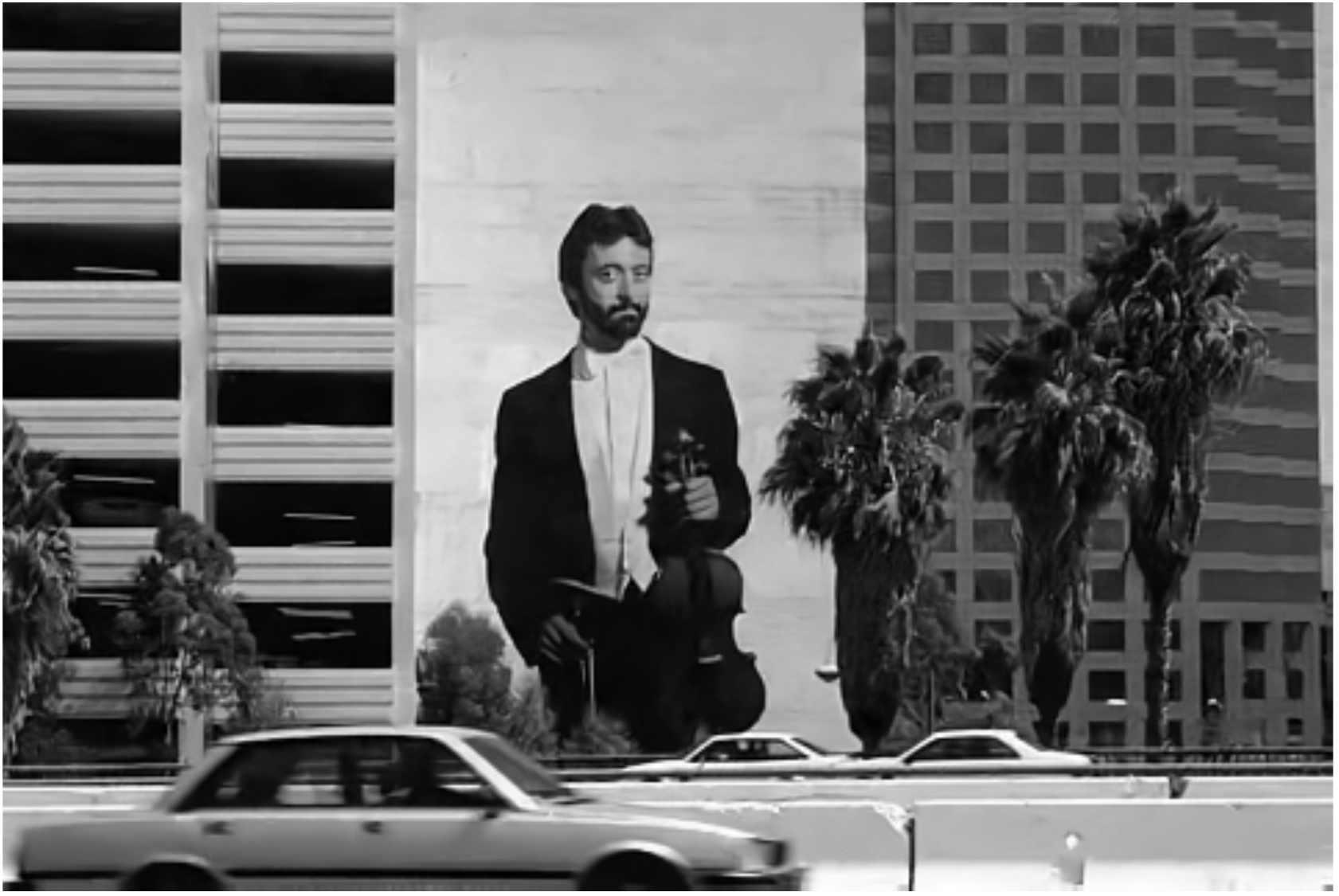}&
\includegraphics[width=.5\linewidth]{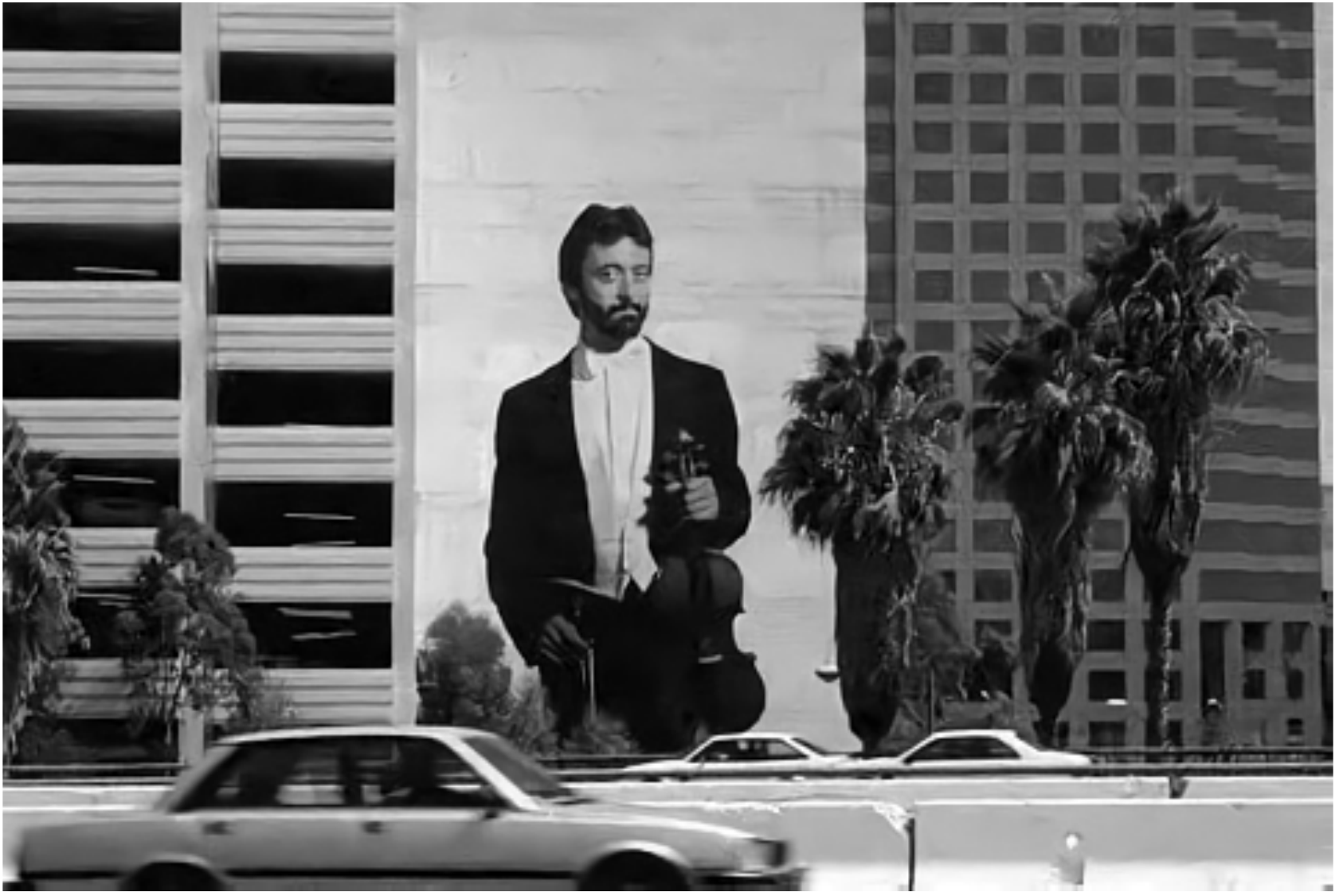}\\
    (c) & (d) \\
\includegraphics[width=.5\linewidth]{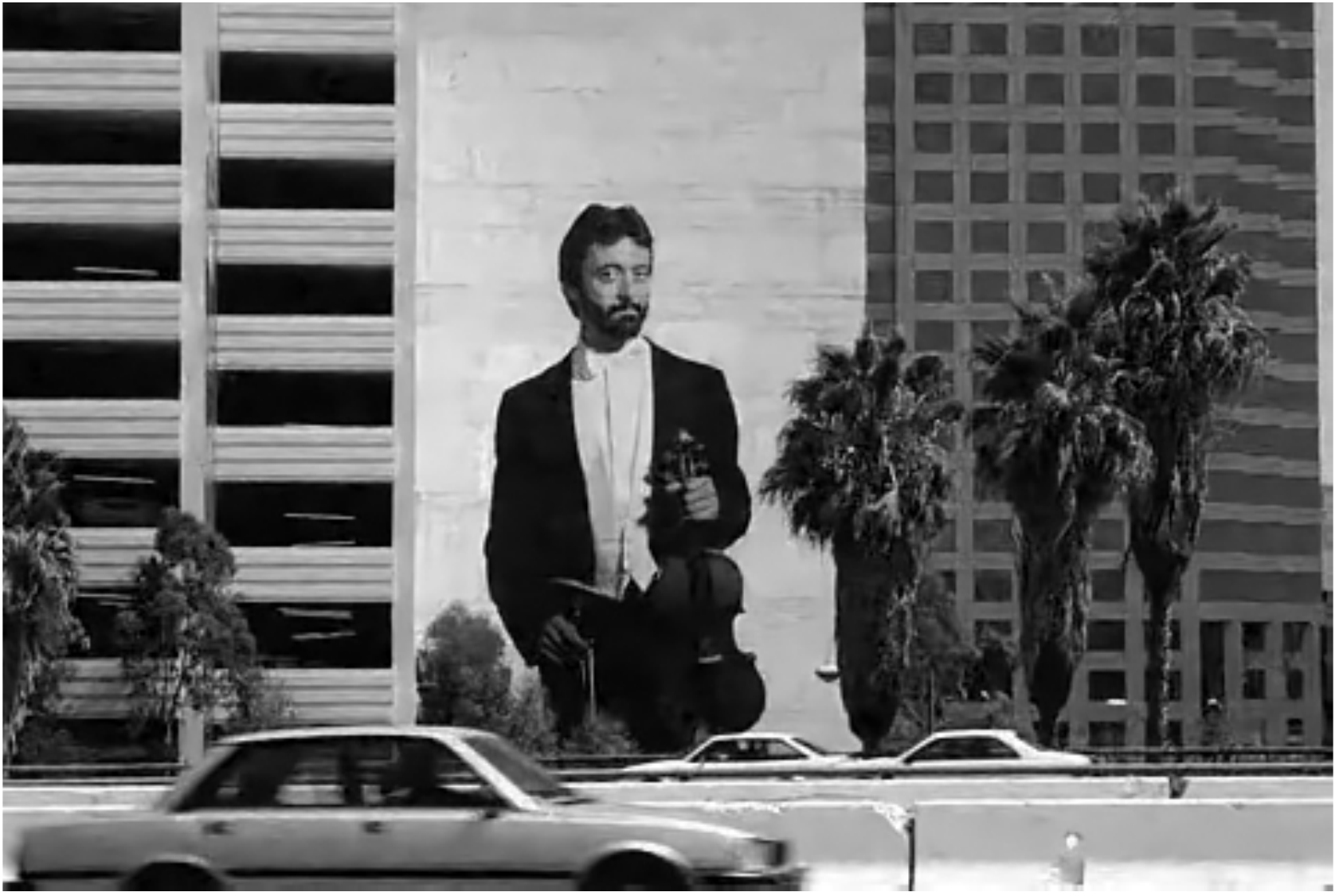}&
\includegraphics[width=.5\linewidth]{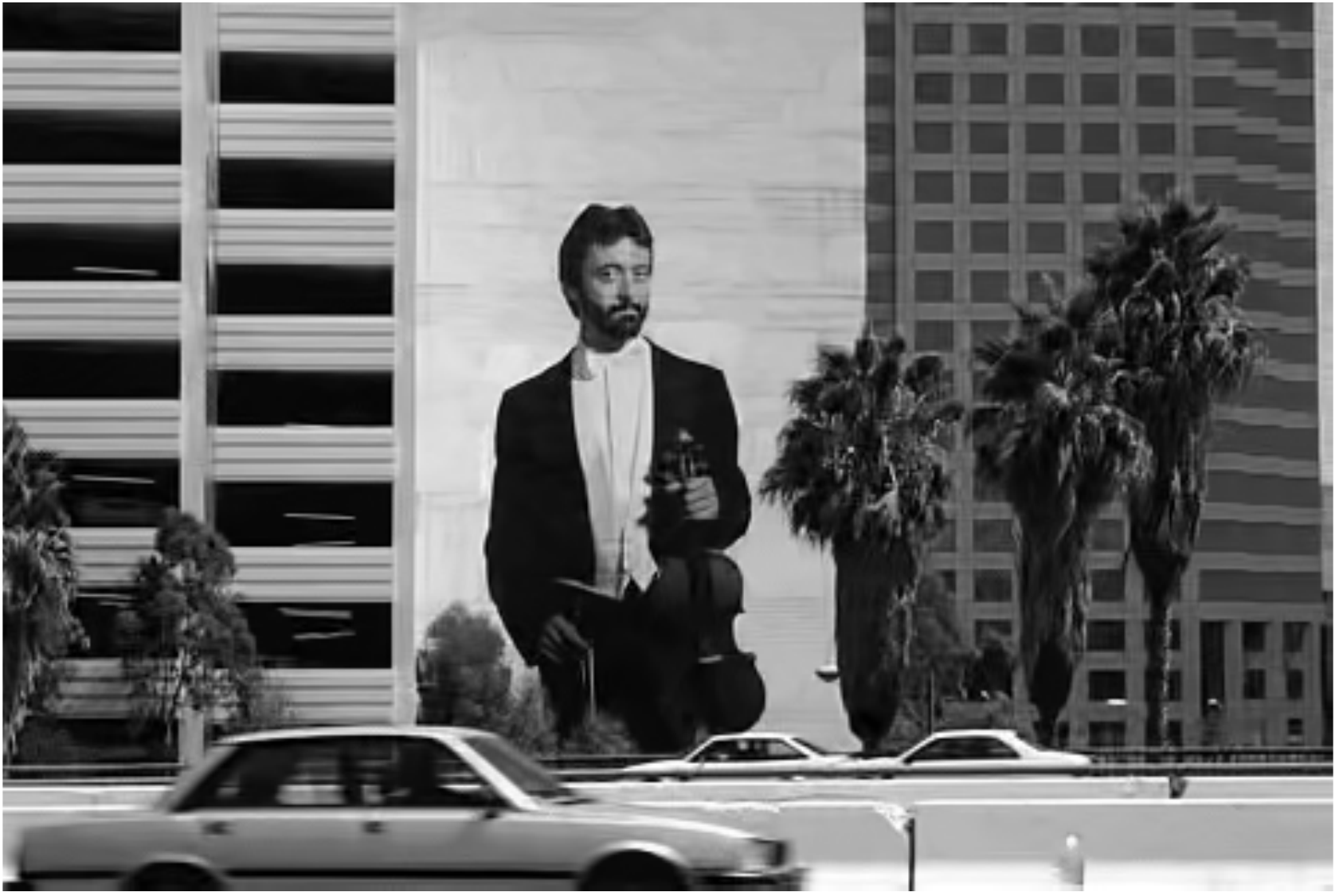}\\
    (e) & (f)    
\end{tabular}
   \caption{Grayscale image denoising. (a) Original image, (b) Noisy image corrupted with Gaussian noise ($\sigma = 15$) ; $\operatorname{PSNR} = 24.57 \text{ dB}$. (c) Denoised image using $\operatorname{NLNet}_{5\times 5}^5$ ; $\operatorname{PSNR} = \textbf{31.58} \textbf{ dB}$. (d) Denoised image using $\operatorname{TNRD}_{7\times 7}^5$~\cite{Chen2016} ; $\operatorname{PSNR} = 31.34 \text{ dB}$. (e) Denoised image using EPLL~\cite{Zoran2011} ; $\operatorname{PSNR} = 31.02 \text{ dB}$. (f) Denoised image using BM3D~\cite{Dabov2007} ; $\operatorname{PSNR} = 31.29 \text{ dB}$. \textbf{Images are best viewed magnified on screen. Note the differences of the denoised results in the highlighted region}.}
   \label{fig:ColorComp}
\end{figure*}

\begin{figure*}[!t]
\centering
\begin{tabular}{@{} c @{ } c @{ } } 
 \includegraphics[width=.5\linewidth]{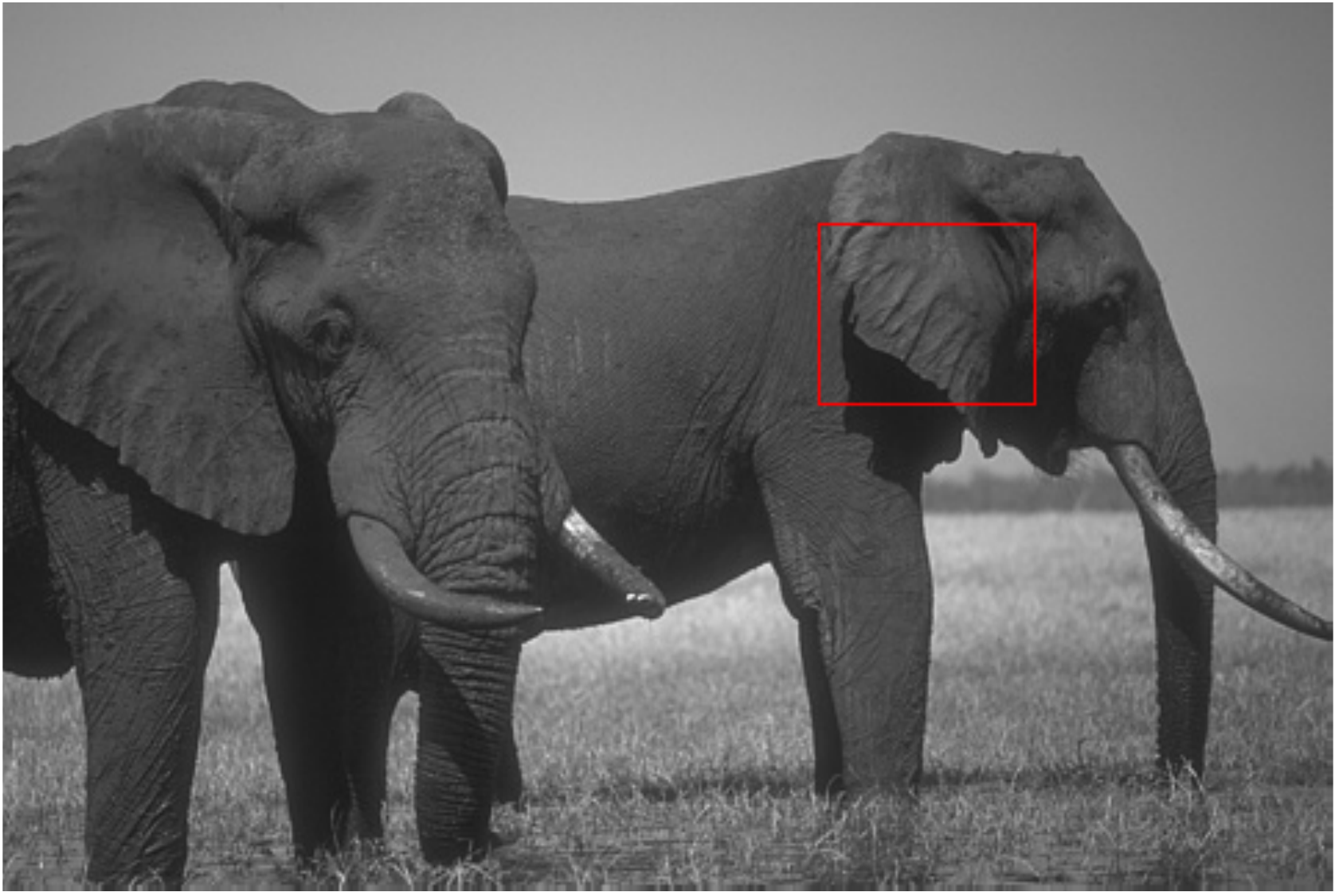}&
\includegraphics[width=.5\linewidth]{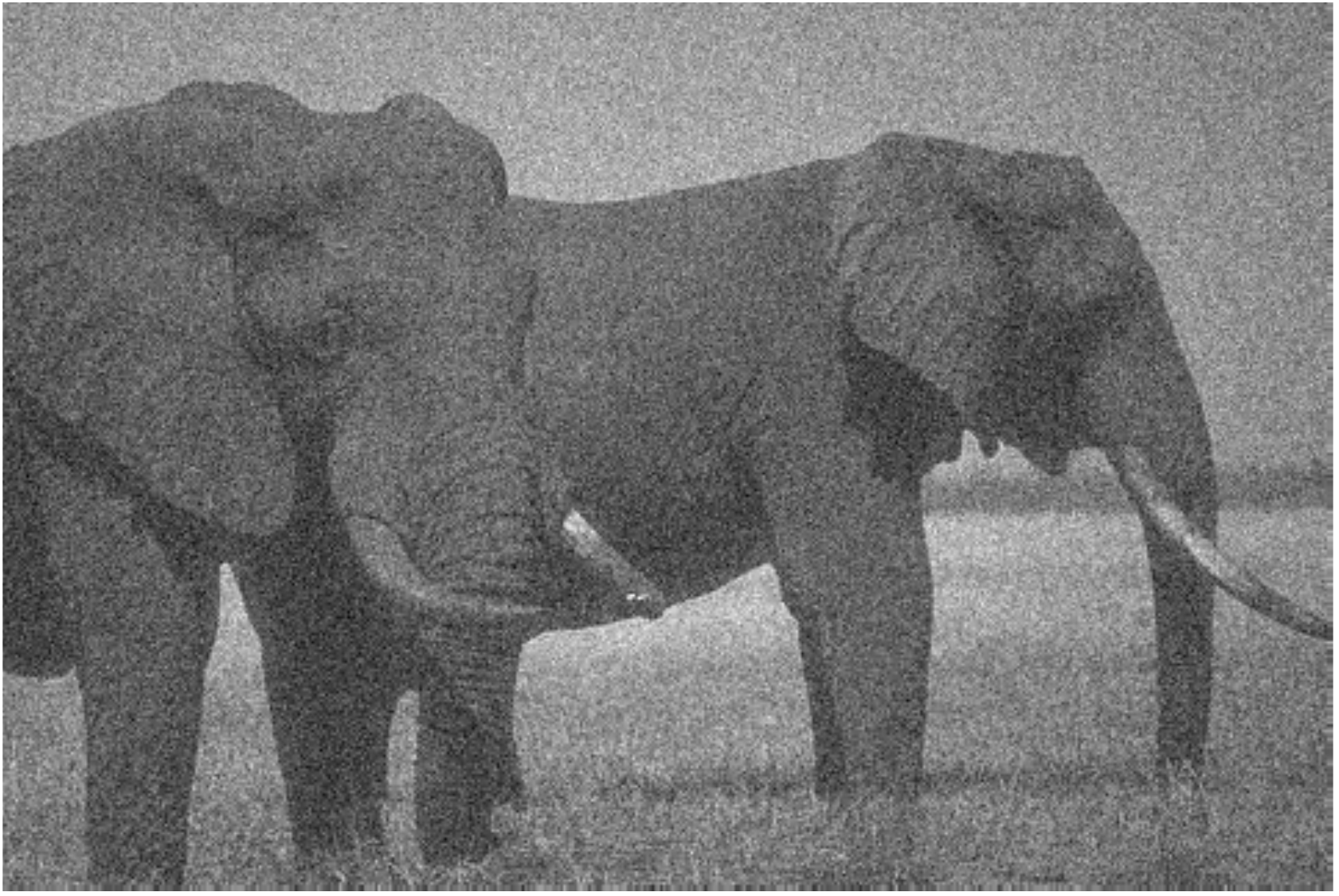}\\
(a) & (b) \\
\includegraphics[width=.5\linewidth]{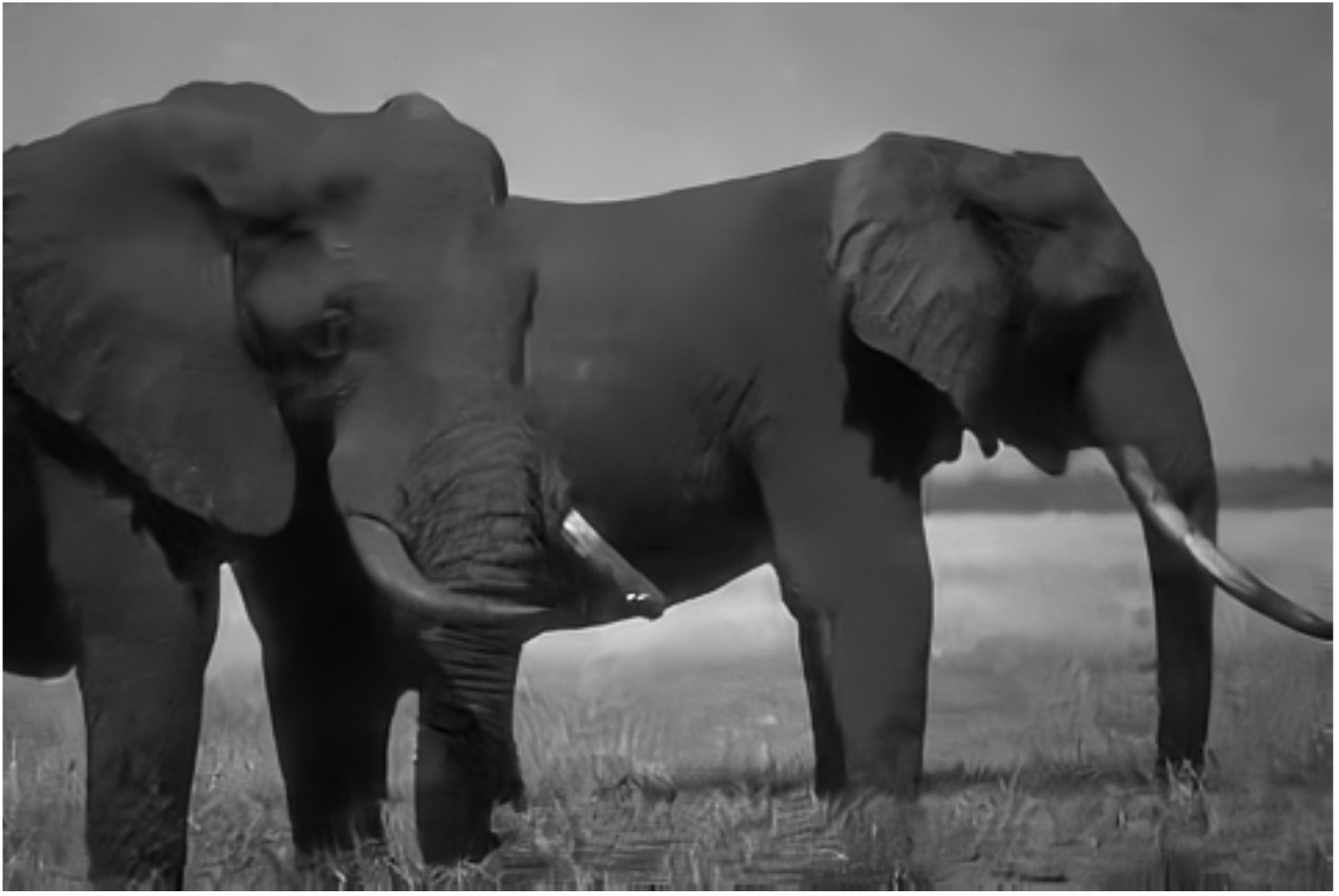}&
\includegraphics[width=.5\linewidth]{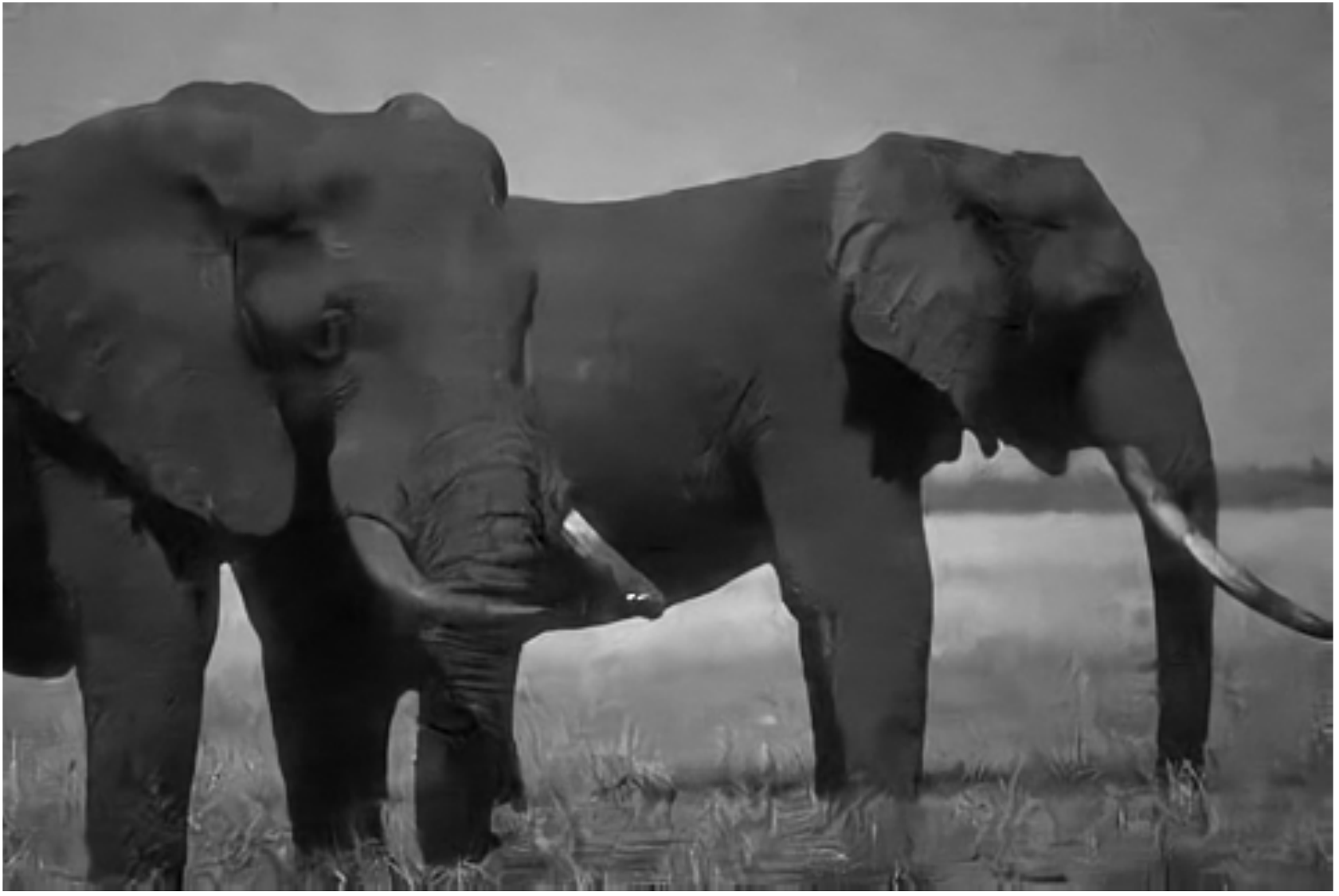}\\
    (c) & (d) \\
\includegraphics[width=.5\linewidth]{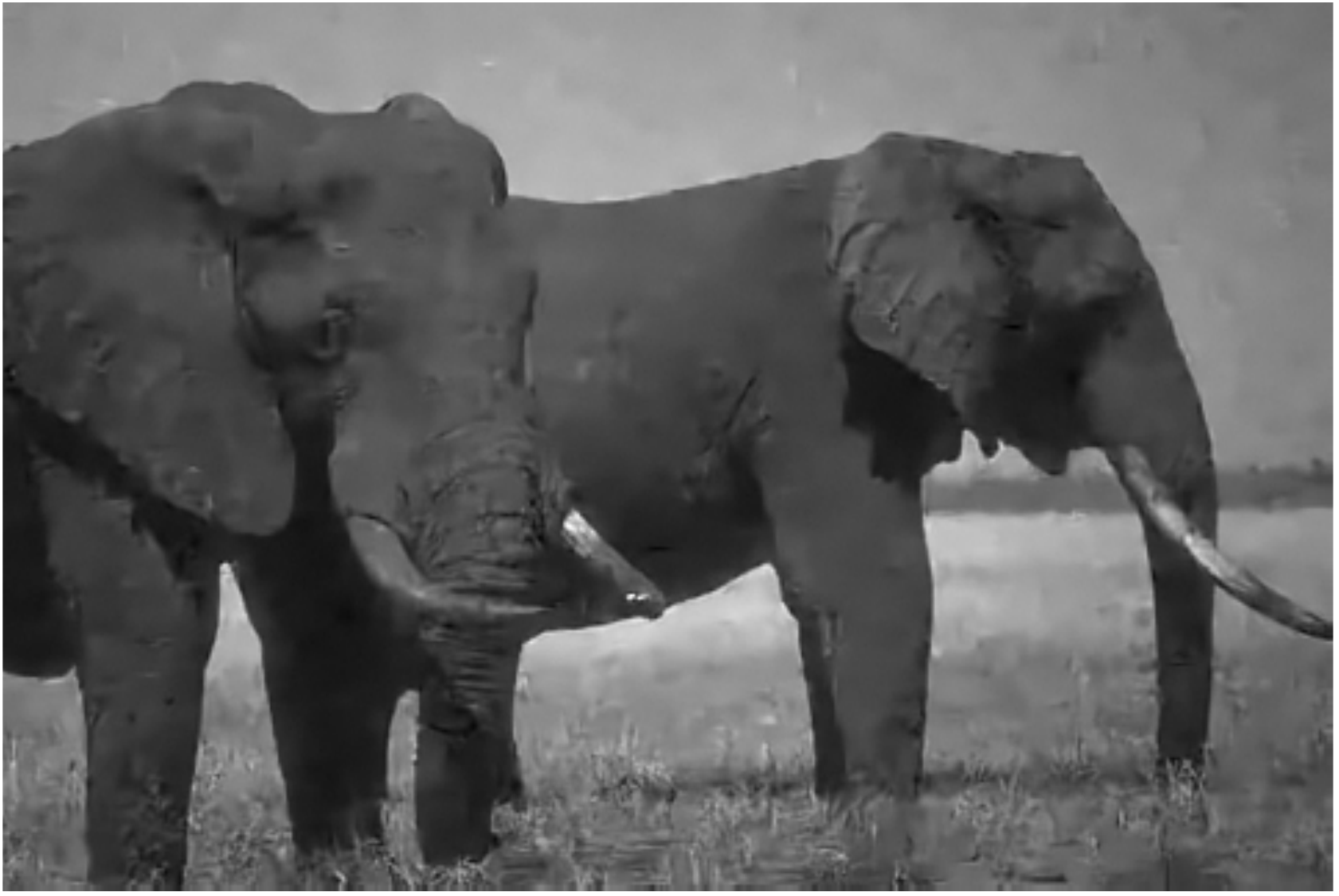}&
\includegraphics[width=.5\linewidth]{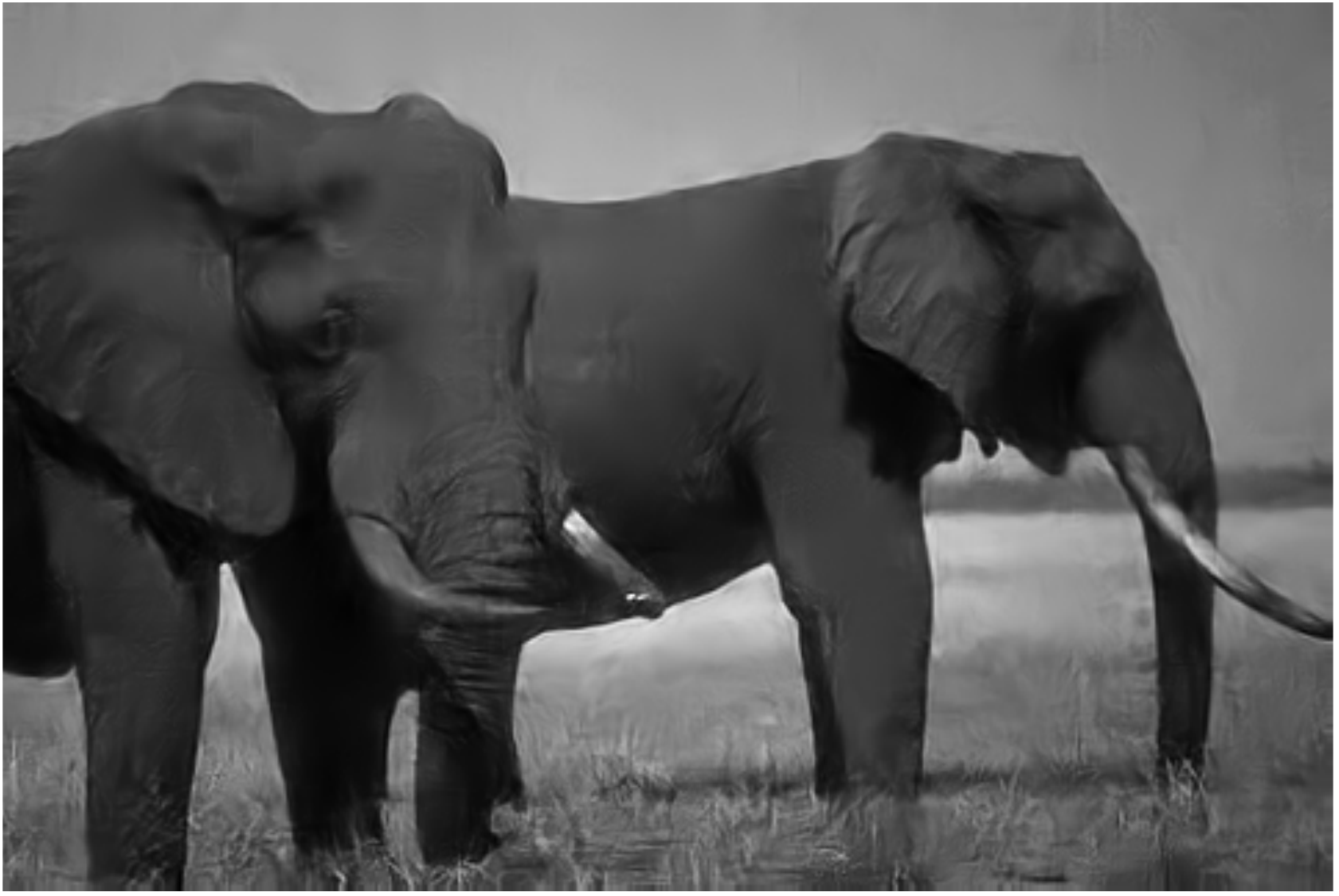}\\
    (e) & (f)    
\end{tabular}
   \caption{Grayscale image denoising. (a) Original image, (b) Noisy image corrupted with Gaussian noise ($\sigma = 25$) ; $\operatorname{PSNR} = 20.18 \text{ dB}$. (c) Denoised image using $\operatorname{NLNet}_{7\times 7}^5$ ; $\operatorname{PSNR} = \textbf{30.25} \textbf{ dB}$. (d) Denoised image using $\operatorname{TNRD}_{7\times 7}^5$~\cite{Chen2016} ; $\operatorname{PSNR} = 30.15 \text{ dB}$. (e) Denoised image using EPLL~\cite{Zoran2011} ; $\operatorname{PSNR} = 29.92 \text{ dB}$. (f) Denoised image using MLP~\cite{Burger2012} ; $\operatorname{PSNR} = 30.16 \text{dB}$. \textbf{Images are best viewed magnified on screen. Note the differences of the denoised results in the highlighted region}.}
   \label{fig:ColorComp}
\end{figure*}

\begin{figure*}[t]
\centering
\begin{tabular}{@{} c @{ } c @{ } c @{ } c @{ } }
 \includegraphics[width=.25\linewidth]{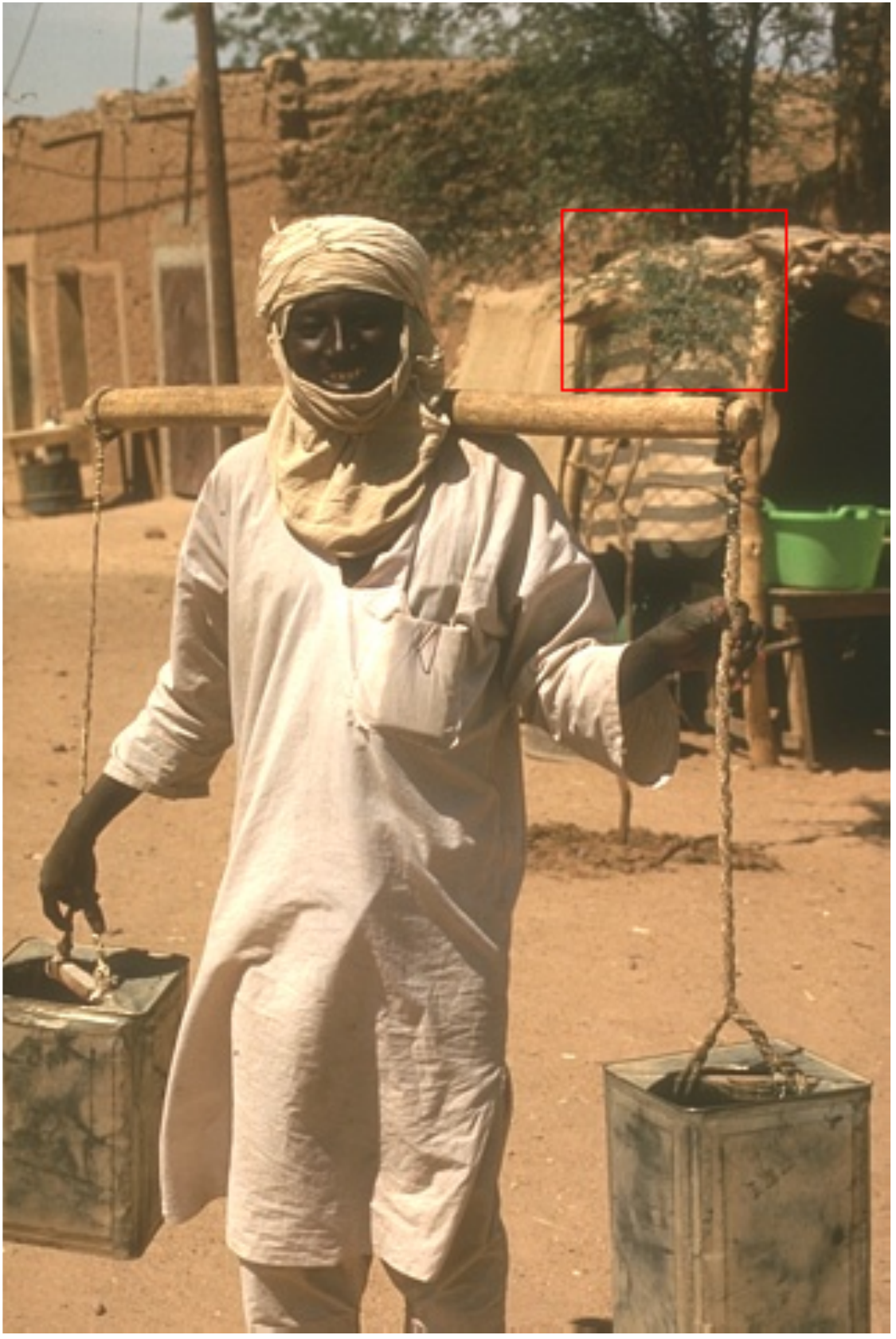}&
 \includegraphics[width=.25\linewidth]{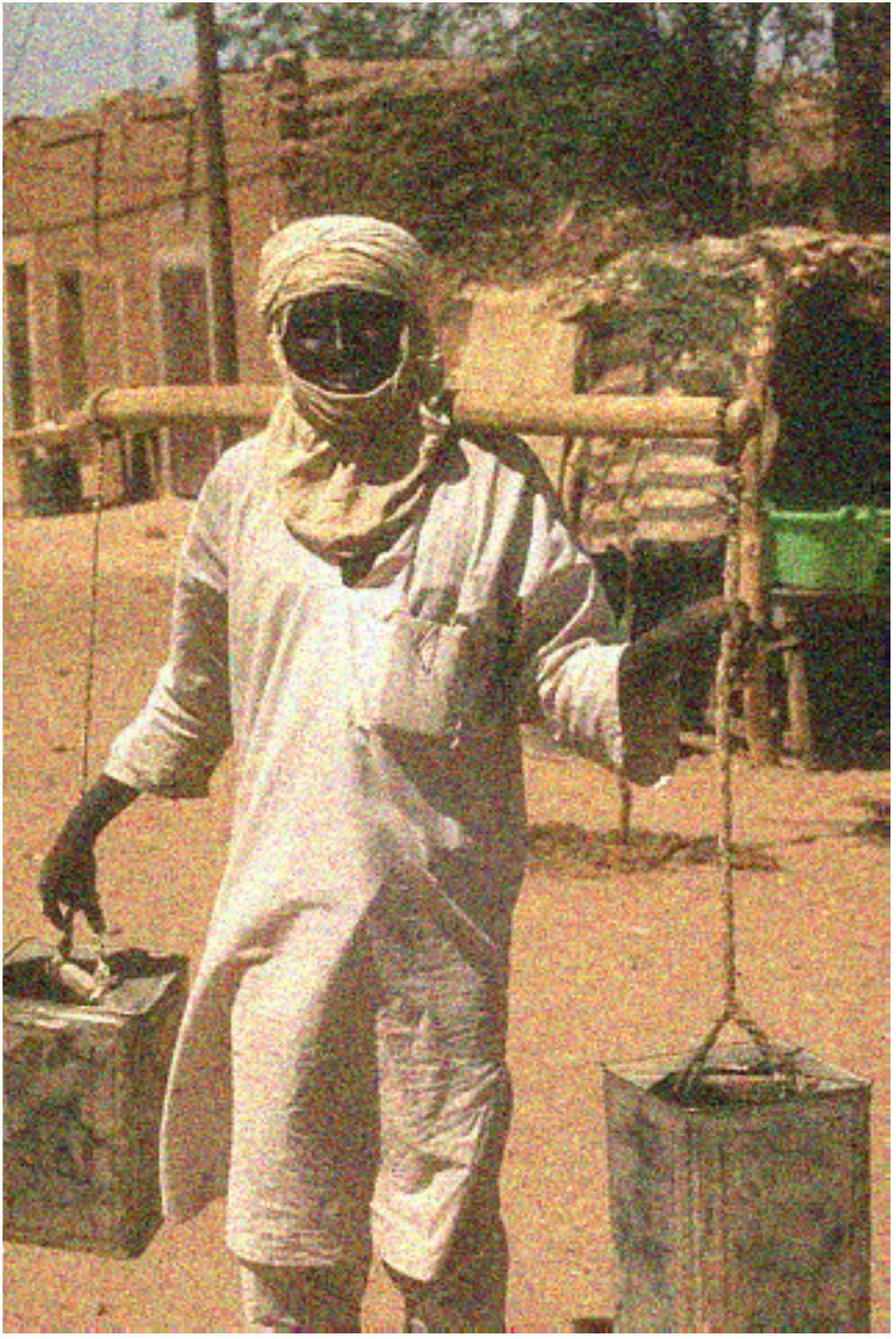}&
 \includegraphics[width=.25\linewidth]{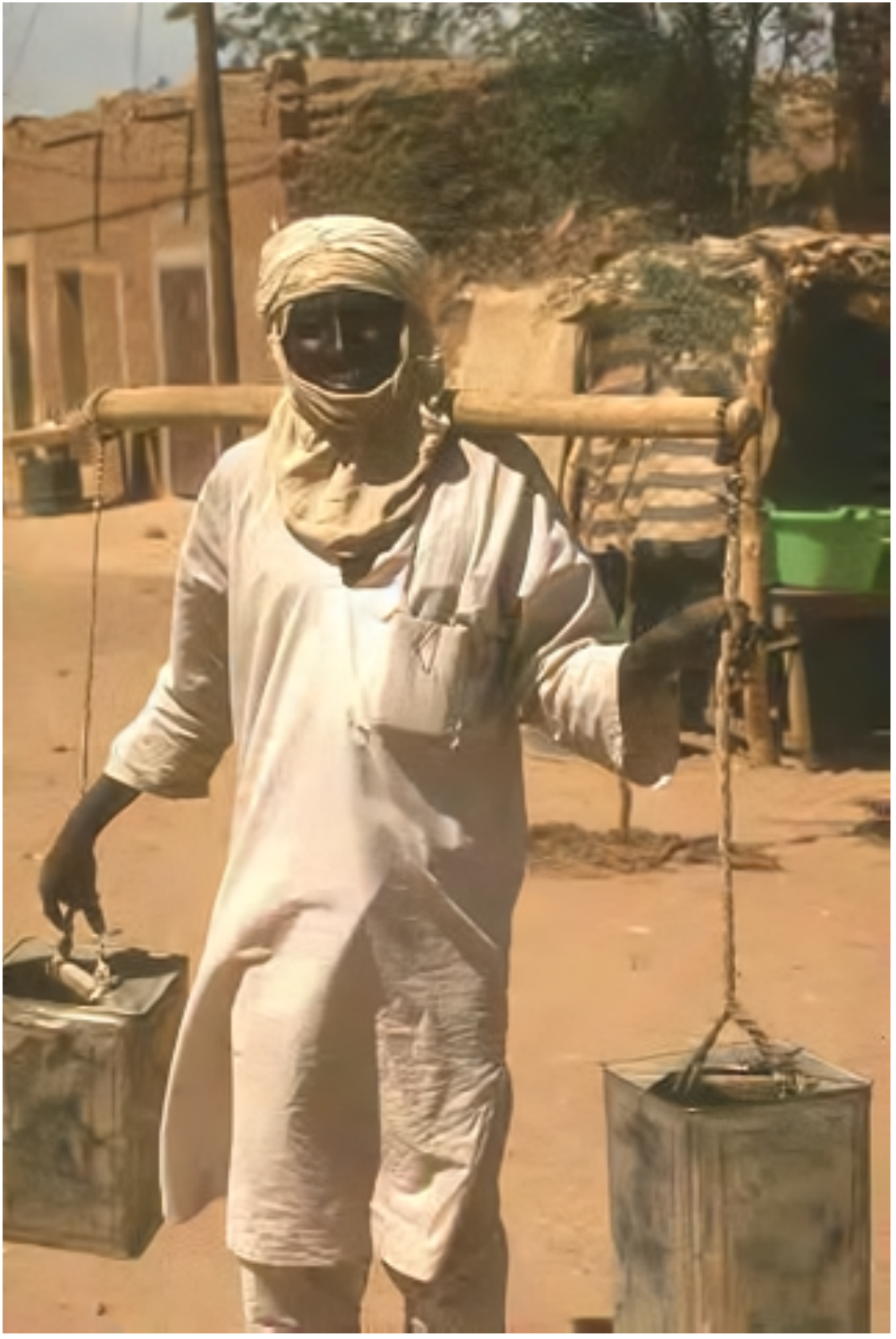}&
 \includegraphics[width=.25\linewidth]{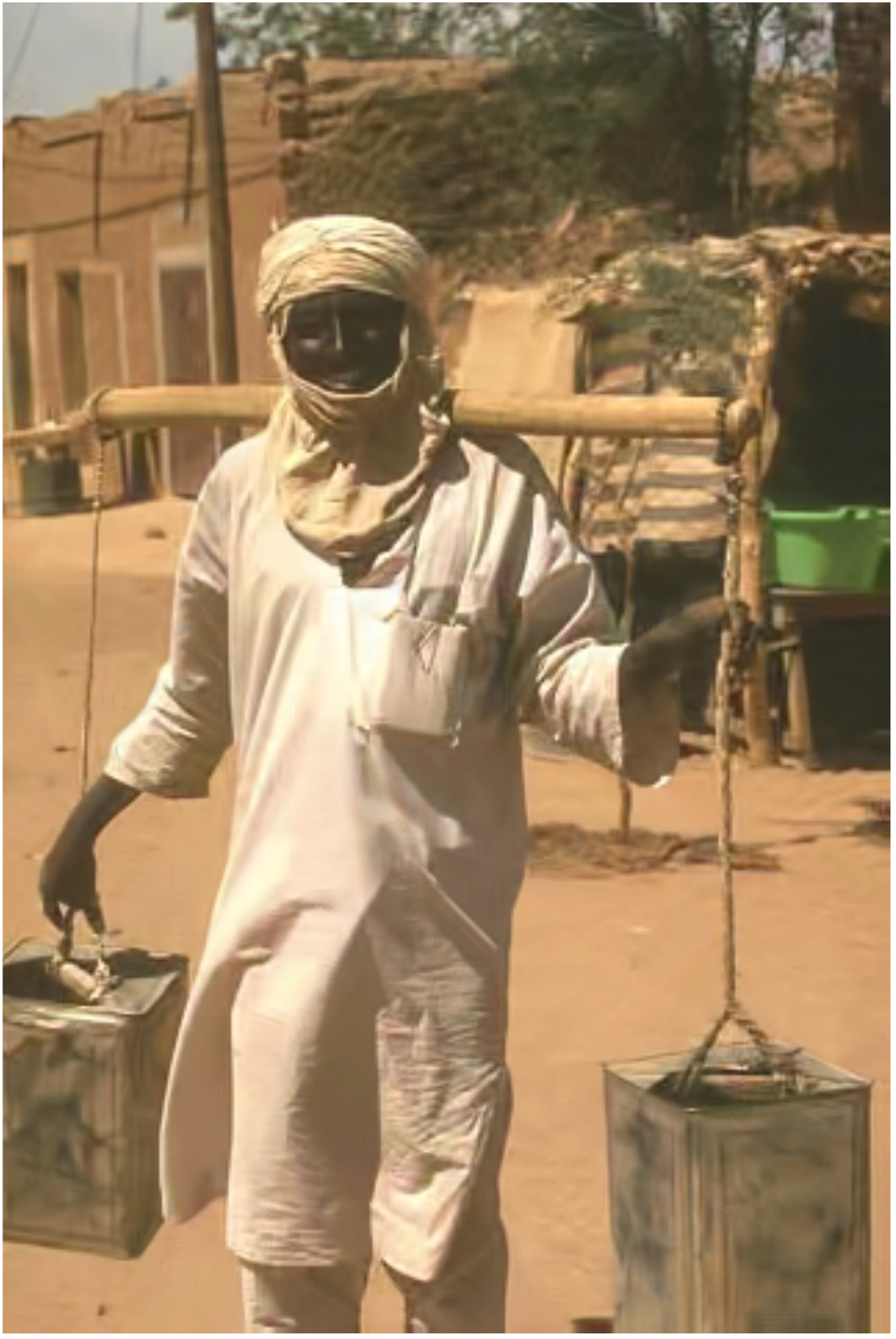}\\
   (a) & (b) & (c) & (d)
\end{tabular}
   \caption{Color image denoising. (a) Original image, (b) Noisy image corrupted with Gaussian noise ($\sigma = 25$) ; $\operatorname{PSNR} = 20.34 \text{ dB}$. (c) Denoised image using $\operatorname{CNLNet}_{5\times 5}^5$ ; $\operatorname{PSNR} = \textbf{31.14} \textbf{ dB}$. (d) Denoised image using  CBM3D~\cite{Dabov2007} ; $\operatorname{PSNR} = 30.75 \text{ dB}$. \textbf{Images are best viewed magnified on screen. Note the differences of the denoised results in the highlighted region}.}
   \label{fig:ColorComp}
\end{figure*}

\begin{figure*}[!t]
\centering
\begin{tabular}{@{} c @{ } c @{ } }
 \includegraphics[width=.5\linewidth]{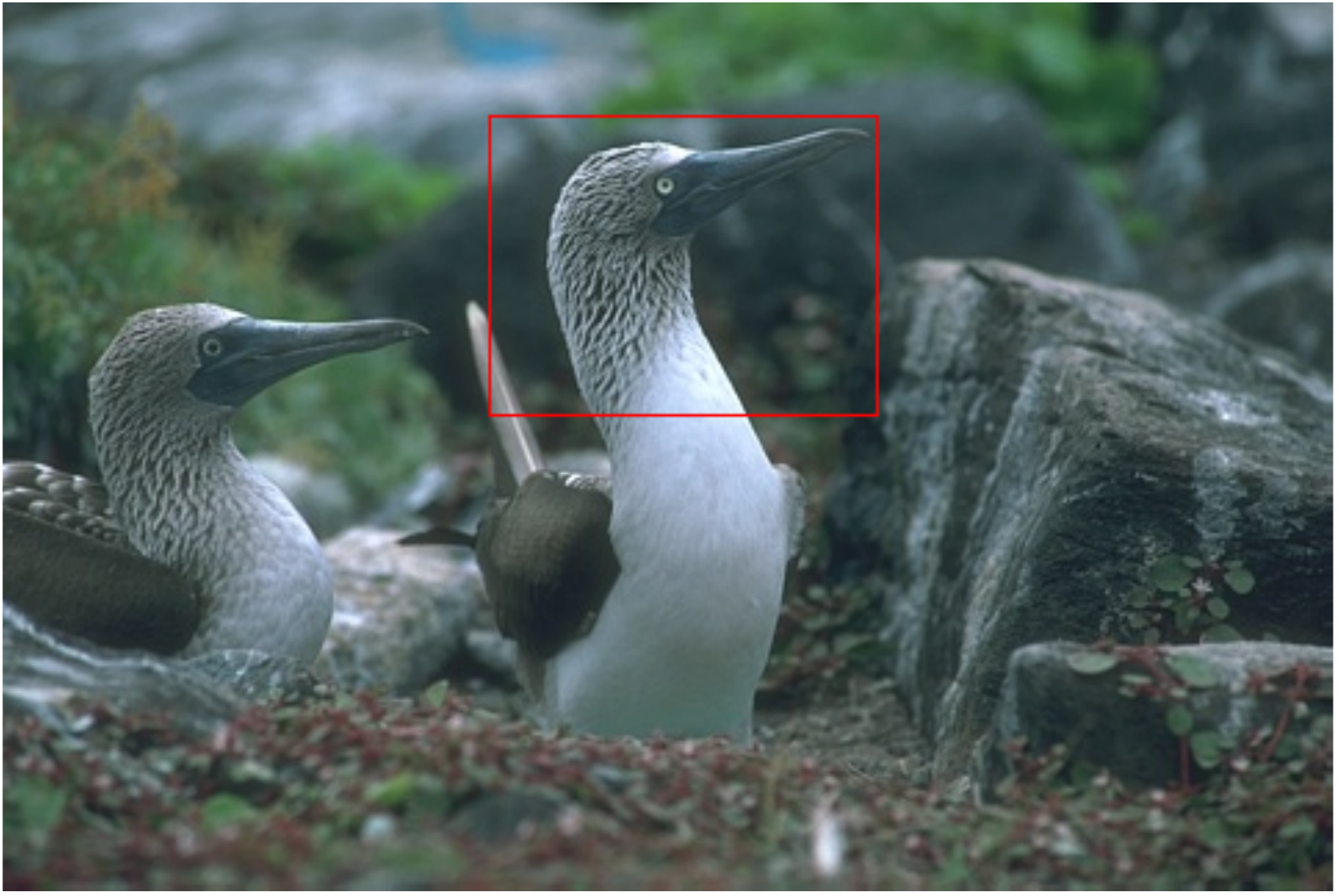}&
 \includegraphics[width=.5\linewidth]{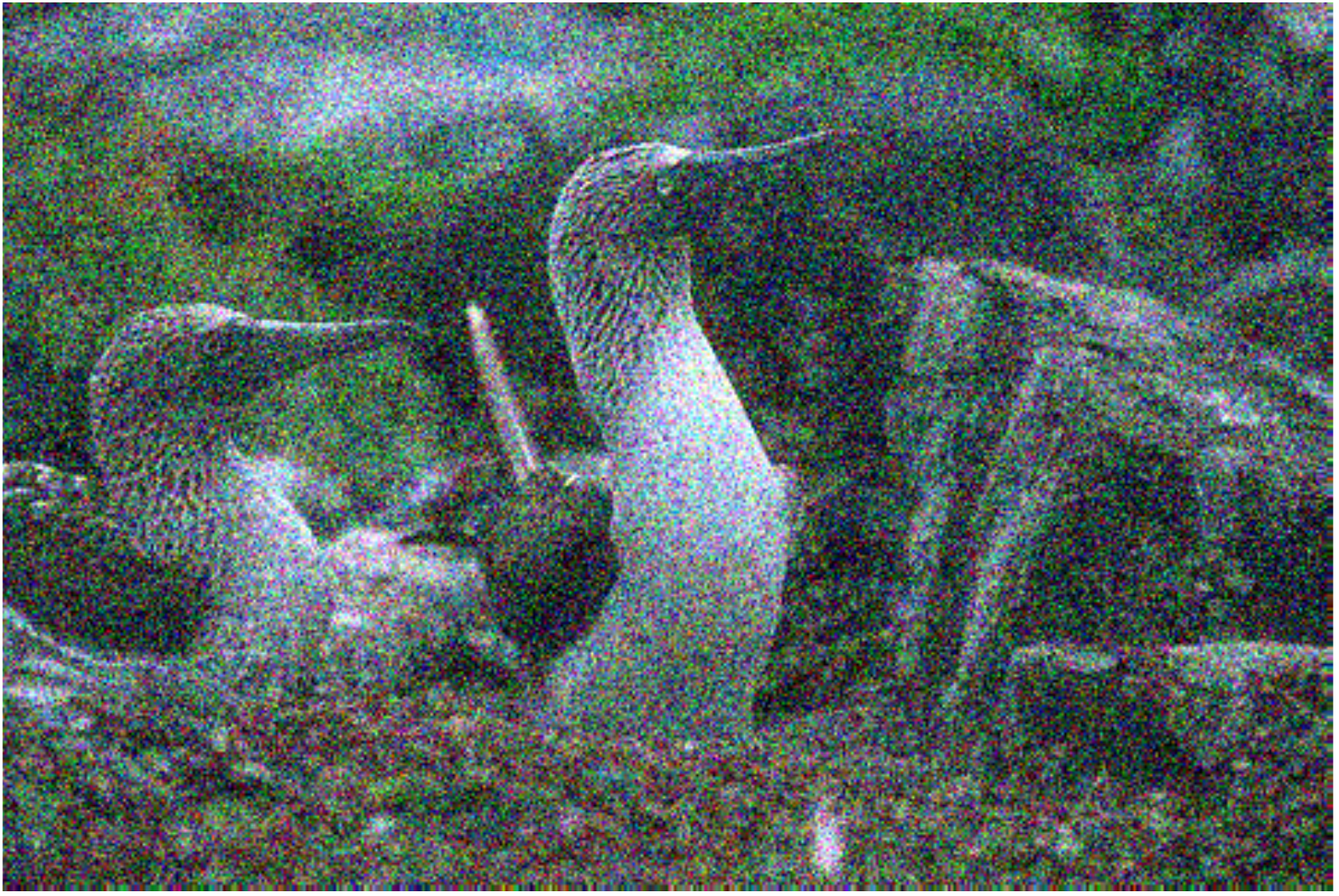}\\
(a) & (b) \\
 \includegraphics[width=.5\linewidth]{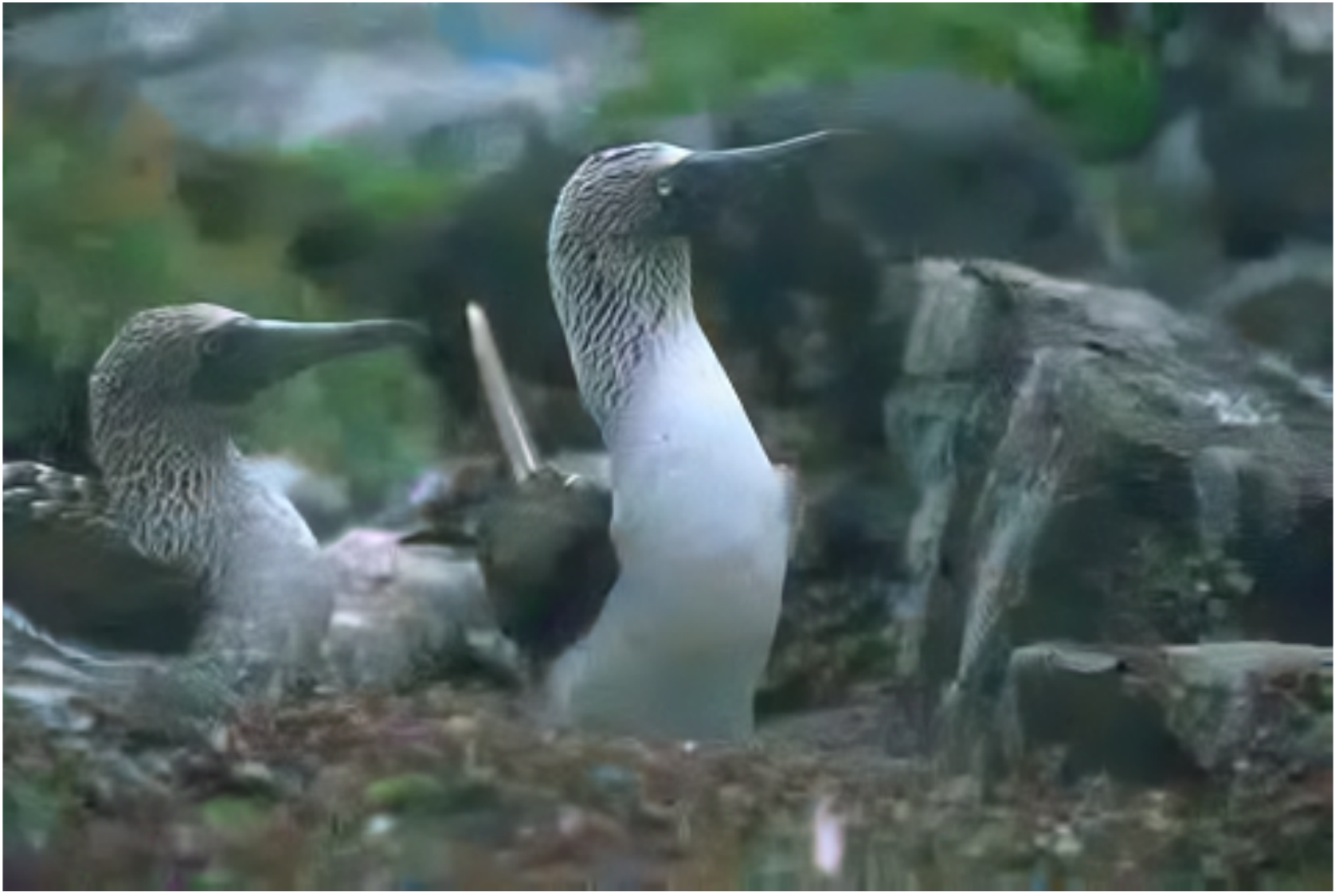}&
 \includegraphics[width=.5\linewidth]{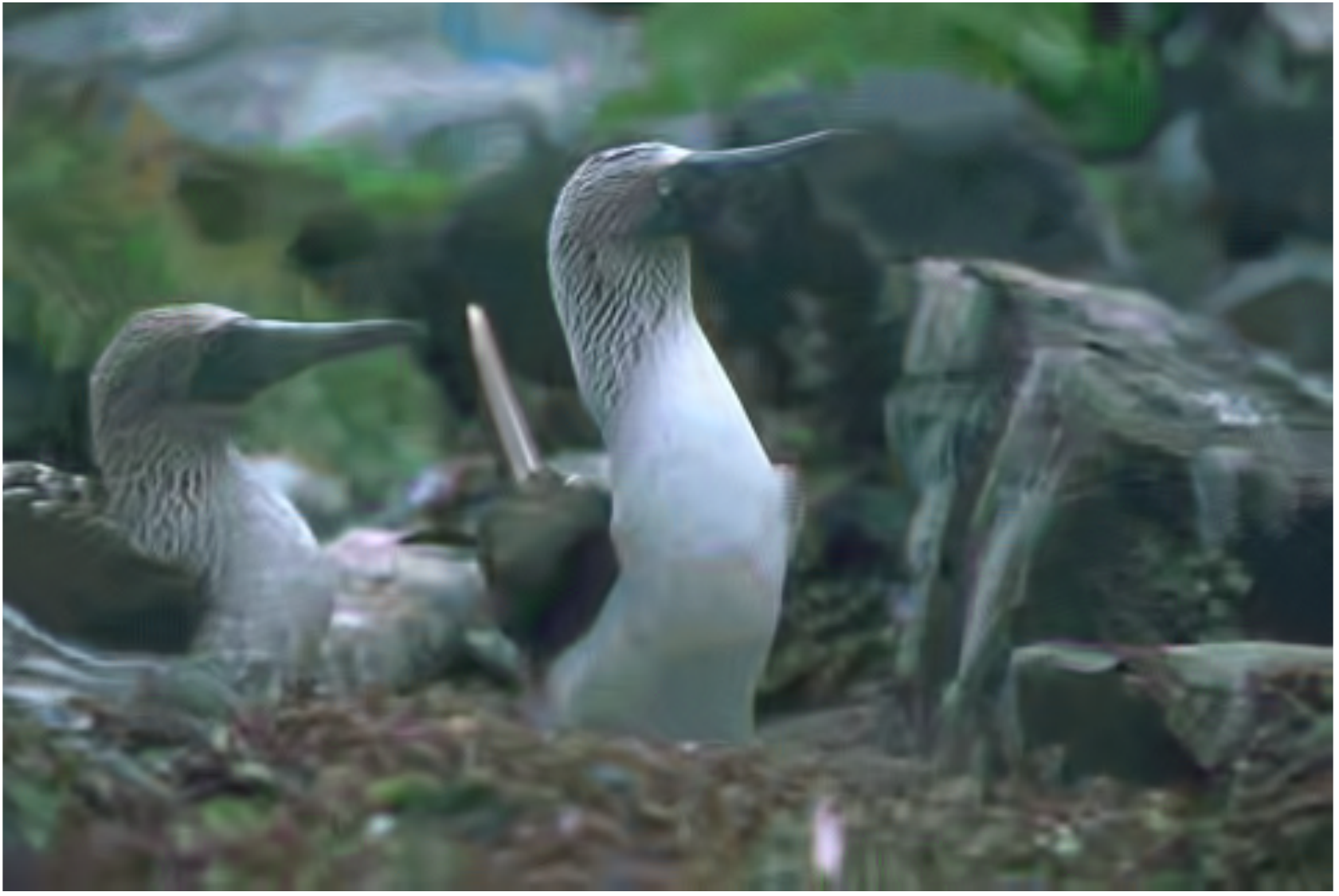}\\
(c) & (d)
\end{tabular}
   \caption{Color image denoising. (a) Original image, (b) Noisy image corrupted with Gaussian noise ($\sigma = 50$) ; $\operatorname{PSNR} = 15.10 \text{ dB}$. (c) Denoised image using $\operatorname{CNLNet}_{5\times 5}^5$ ; $\operatorname{PSNR} = \textbf{24.74} \textbf{ dB}$. (d) Denoised image using  CBM3D~\cite{Dabov2007} ; $\operatorname{PSNR} = 24.39 \text{ dB}$. \textbf{Images are best viewed magnified on screen. Note the differences of the denoised results in the highlighted region}.}
   \label{fig:ColorComp}
\end{figure*}

\section{Acknowledgments}
The author would like to thank NVIDIA for supporting this work by donating a Tesla K-40 GPU.

{\small
\bibliographystyle{ieee}
\bibliography{references}
}

\end{document}